\pdfoutput=1

\documentclass[11pt]{article}

\usepackage{ACL2023}

\usepackage{times}
\usepackage{latexsym}
\usepackage{longtable} 
\usepackage{enumitem}
\usepackage{tcolorbox}
\usepackage{booktabs}  
\usepackage{array}     
\usepackage{geometry}  
\usepackage{setspace}  
\usepackage{graphicx}
\usepackage{subcaption}
\usepackage{float}
\usepackage{adjustbox}
\renewcommand{\arraystretch}{1.5} 
\usepackage[T1]{fontenc}

\usepackage[utf8]{inputenc}

\usepackage[utf8]{inputenc}
\usepackage{amsmath}
\usepackage{amsfonts}
\usepackage{graphicx}
\usepackage{inconsolata}
\usepackage{xcolor}
\usepackage{microtype}

\title{Concept-Based Interpretability for Toxicity Detection}

\author{
  Samarth Garg \\
  ABV–IIITM, Gwalior, India \\
  \texttt{samarthgarg0904@gmail.com}
  \And
  Divya Singh \\
  IIT Patna, India \\
  \texttt{divya.24aug@gmail.com}
  \AND
  Deeksha Varshney \\
  IIT Jodhpur, India \\
  \texttt{deeksha@iitj.ac.in}
  \And
  Mamta \\
  IIT Patna, India \\
  \texttt{1921cs11@iitp.ac.in}
}

\begin{document}
\maketitle

\begin{abstract}
The rise of social networks has not only facilitated communication but also allowed the spread of harmful content. Although significant advances have been made in detecting toxic language in textual data, the exploration of concept-based explanations in toxicity detection remains limited. In this study, we leverage various subtype attributes present in toxicity detection datasets, such as \textit{obscene, threat, insult, identity attack, and sexual explicit} as concepts that serve as strong indicators to identify whether language is toxic. However, disproportionate attribution of concepts towards the target class often results in classification errors. Our work introduces an interpretability technique based on the Concept Gradient (CG) method which provides a more causal interpretation by measuring how changes in concepts directly affect the output of the model. This is an extension of traditional gradient-based methods in machine learning, which often focus solely on input features. We propose the curation of \textit{Targeted Lexicon Set}, which captures toxic words that contribute to misclassifications in text classification models. To assess the significance of these lexicon sets in misclassification, we compute Word-Concept Alignment (WCA) scores, which quantify the extent to which these words lead to errors due to over-attribution to toxic concepts. Finally,  we introduce a lexicon-free augmentation strategy by generating toxic samples that exclude predefined toxic lexicon sets. This approach allows us to examine whether over-attribution persists when explicit lexical overlap is removed, providing insights into the model's attribution on broader toxic language patterns.
\end{abstract}

\section{Introduction}
The increasing use of social media has made communication more accessible, but it has also facilitated the spread of questionable and harmful content, which can sometimes result in real-world consequences, such as criminal activities \citep{doi:10.1080/1369118X.2017.1293130,johnson2019hidden,home2017hate,center2017year}. To safeguard individuals' rights to free expression while maintaining the safety of online spaces, many platforms implement automated content moderation systems using pre-trained language models \citep{devlin-etal-2019-bert,liu2019roberta} to detect and flag content that breaches community guidelines \citep{10.1145/2740908.2742760,10.1145/2872427.2883062,macavaney:plosone2019-hate,cao2023toxicity,gururangan-etal-2020-dont,sotudeh-etal-2020-guir,wiedemann-etal-2020-uhh,zampieri-etal-2020-semeval}.

A significant challenge in deploying these systems lies in their decision-making opacity, making it difficult to explain the predictions of the models. However, interpretable models provide transparency, assisting moderators in reviewing flagged content more efficiently, and fostering trust through clear explanations of prediction mechanisms. Although various feature attribution methods have been developed to identify important input features for model decisions \citep{xiang2021toxccin,han2022hierarchical,zhang2023neurosymbolic}, these approaches are often limited when the features are not intuitive to human understanding. This shortcoming highlights the need for concept-based explanations, where models generate explanations based on human-understandable concepts \citep{nejadgholi2022improving,nejadgholi2023concept,ge2022explainable}. However, disproportionate attribution of certain human-defined concepts \citep{kim2018interpretability,nejadgholi2023concept} towards target class often results in classification errors. 

Large Language Models (LLMs) have proven to be highly effective in tasks such as emotion and sentiment analysis due to their ability to grasp nuanced linguistic features \citep{batra2021bert,biswas2020achieving}. Models like ChatGPT, often deployed in "zero-shot" settings without task-specific fine-tuning, excel in many areas but face challenges when detecting implicit forms of toxicity, especially in cases involving figurative language, such as metaphor, sarcasm or irony \citep{mishra2024exploring}. Consider a content moderation system employed on a social media platform to identify toxic comments. A content moderator may question why the model flagged a specific comment as toxic. Did the model determine that the presence of certain words, like "hate" or "stupid," was sufficient to classify the comment as harmful? Furthermore, the moderator might want to intervene in the model’s decision-making process: if they were to inform the model that the comment was actually a critique of toxic behavior rather than promoting it, would the model revise its toxicity assessment?

In this work, we propose a novel interpretability method based on Concept Gradients (CG) to address the challenges of understanding model behavior in toxicity detection. The novelty of our method lies in its ability to measure how changes in specific toxicity-related concepts directly affect model predictions, providing a more causal and transparent explanation compared to traditional gradient-based approaches. To enhance the robustness of our findings, we also propose a lexicon-free augmentation strategy, which tests whether the over-reliance on specific toxic lexicons persists when such terms are excluded from the training data. We evaluate our method on the task of toxicity detection using multiple benchmark datasets, including the Civil Comments and HateXplain datasets. The evaluation results show that our method improves model interpretability and helps identify potential biases in toxicity classification.




Our contribution can be summarized as follows: 1) We propose the creation of a \textit{Targeted Lexicon Set} that identifies and captures toxic words contributing to misclassifications using Concept Gradient scores \citep{bai2023concept}, facilitating a systematic evaluation of model behavior in relation to specific toxic subtypes. 2) We compute Word-Concept Alignment (WCA) scores to assess the significance of the curated lexicon sets, quantifying the extent to which specific toxic words lead to misclassifications due to over-attribution to toxic concepts. 3) We present a lexicon-free augmentation strategy that generates toxic samples without predefined toxic lexicons, allowing us to explore the persistence of over-attribution in the absence of explicit lexical overlap and providing deeper insights into broader toxic language patterns\footnote{Data and source code will be released upon acceptance.}.



\section{Related Works}
Addressing negativity in online spaces is vital, as the prevalence of digital conversations necessitates effective automated toxicity detection systems to foster constructive discussions and enhance the overall online experience.  Earlier studies predominantly focused on surface-level features, such as bag-of-words models, along with traditional machine learning techniques \citep{warner-hirschberg-2012-detecting, waseem-hovy-2016-hateful, davidson2017automated}. While these methods have been shown to be quite predictive \citep{schmidt-wiegand-2017-survey} and straightforward to interpret, they often face issues with false positives, where specific patterns can lead to misclassifications \citep{kwok2013locate}.  For instance, some slurs commonly found in African American English may be misinterpreted as clear indicators of toxicity \citep{xia-etal-2020-demoting}, despite being innocuous terms used within the Black online community. These features also require such predictive terms to appear in both training and testing set to work effectively. 

Neural text representations have proven effective in detecting toxicity \citep{djuric2015hate,nobata2016abusive}, with large pre-trained models like BERT \citep{devlin-etal-2019-bert} showing advantages by using contextual word embeddings \citep{zampieri2019semeval, zampieri-etal-2020-semeval}. Given the implications of misclassifications in toxic language detection, interpretability is crucial for transparency in decision-making. One approach is providing explanations, such as rationales, which highlight significant sub-strings affecting model predictions \citep{belinkov-glass-2019-analysis, zaidan-etal-2007-using}. For example, \citep{zhang-etal-2016-rationale} introduced a CNN model using rationales, and \citep{mathew2021hatexplain} developed a hate speech dataset with rationales and labels, though their baseline models were primarily for dataset evaluation. Concept-based explanations have been primarily used in computer vision for image classification \citep{graziani2018regression, ghorbani2019towards, yeh2020completeness} and in NLP to evaluate abusive language classifiers for COVID-related anti-Asian hate speech \citep{nejadgholi2022improving}, sentiment fairness \citep{nejadgholi2023concept}, and topic modeling-based classifiers \citep{yeh2020completeness}. In \citep{nejadgholi2022improving}, a TCAV-based interpretability technique and the Degree of Explicitness metric enhance detection of implicit abuse. \citep{mishra2024exploring} analyze ChatGPT's toxicity detection on GitHub comments, uncovering common errors. While previous research has primarily treated concepts as linearly seperable entities, our approach distinguishes itself by acknowledging and exploring the complex, non-linear relationships between concepts. To the best of our knowledge, we are the first to investigate this intricate interplay, which allows for a deeper understanding of how concepts interact and influence model behavior. This perspective marks a significant advancement in the field, paving the way for more nuanced interpretations of model predictions.

Data augmentation \cite{dixon2018measuring,badjatiya2019stereotypical,han2020fortifying,waseem2018bridging,karan2018cross} has been employed to enhance the robustness of classifiers that detect abusive content. In contrast to these methodologies, our approach emphasizes utilizing data augmentation as a complementary strategy for analyzing the over-attribution of model predictions.


\section{Methodology}

\subsection{Problem Definition}
Let the set of samples be denoted by $ \mathcal{D} = \{(x_i, y_i)\}_{i=1}^{N}$, where $x_i$ is the input sentence and $y_i \in \{0, 1\}$ is the corresponding binary label indicating non-toxic (0) or toxic (1).  The primary objective is to interpret the intermediate activations of a neural network by utilizing a set of predefined concepts, $C$ = $\{c_1, . . . , c_m\}$. The task involves constructing a relationship between these concepts and model predictions to explain the behavior of the toxicity prediction model. The secondary task focuses on constructing \textit{Targeted Lexicon Sets} ($T_i$), each containing specific words $w_j$ that are highly relevant to misclassifications, to understand which lexical items contribute most to the model's errors in toxicity prediction.

\subsection{Concept-based Interpretation}
\label{sec:cg_compute}
For the toxicity detection task, we take the assumption of non-linearity in concepts, since toxic language is inherently complex and context-dependent.
\citep{bai2023concept} proposed gradient-based method  called Concept Gradients (CG), for identifying which concepts explain model's predictions and it deals with concepts that cannot be expressed as linear combinations of input (or latent) features.

We define a systematic approach for implementing Concept Gradients (CG) to interpret the influence of concepts on the toxicity prediction model. First, we define the model for predicting toxicity as our target model to be interpreted, denoted as \( f \). The choice of the target model is crucial, as it dictates the baseline against which the relevance of the concepts will be assessed. Next, we establish the concept model, referred to as \( g \), to represent the $C$, concepts of interest, which in our case includes various subattributes associated with toxic language. The concept model g takes input embeddings x and outputs activations corresponding to the predefined set of concepts $C$. This flexibility allows for a more customized approach that accommodates the specific nuances present in the dataset.

Once both models are defined, adapted from \citep{bai2023concept}, the concept model, \( g \) is trained using pretrained weights from the target model \( f \) to ensure that both models share a similar architectural foundation and weight configuration. Following the training of \( g \), we compute the concept gradients, $CG(x, f, g)$, 
for each input sample \( x \). This involves calculating the gradients \( \nabla g(x) \) and \( \nabla f(x) \), which reflect how changes in the concept model affect the predictions of the target model.
\begin{equation*}
\label{eq:cg}
    CG(x, f, g) := {\nabla g(x)^\dagger \nabla f(x)},
\end{equation*}
where \( \nabla g(x) \) is the gradient of the concept function \( g \) with respect to input \( x \); \( \nabla f(x) \) is the gradient of the target function \( f \) with respect to input \( x \); \( \nabla g(x)^\dagger \) denotes the Moore–Penrose pseudo-inverse of \( \nabla g(x) \), where $\nabla g(x)^\dagger = \frac{\nabla g(x)^T}{\|\nabla g(x)\|^2}$. The normalization ensures that CG captures the directional influence of \( g \) on \( f \) independently of gradient magnitudes.

Instead of simply counting positive relevance as shown in \citep{bai2023concept}, we calculate the mean absolute value of the local relevance scores across all samples $(x_i,y)$ for a class $y$ denoted as $\text{MeanCG}(X_y; f, g)$. This method accounts for both positive and negative relevance contributions.
\[
\text{MeanCG}(X_y; f, g) = \frac{1}{|X_y|} \sum_{(x, y) \in X_y} \left| \text{CG}(x; f, g) \right|,
\]
where $X_y = \{x_i \mid (x_i, y) \in D\}$.




\subsection{Targeted Lexicon Set Curation}
\label{sec:tlsc}
Following the computation of Concept Gradients (Section \ref{sec:cg_compute}), we aim to create {Targeted Lexicon Set}, $T_i$, that represents the key words that contribute to misclassifications. First, we identify the set of misclassified samples \( \mathcal{D}_{mis} \) from the test dataset, \( \mathcal{D}_{test} \), as \(\mathcal{D}_{mis} = \{x \in \mathcal{D}_{test} \mid y_j \neq \hat{y}_j \} \). Here, \(\hat{y}_j\) denotes the predicted label from the target model for a given input sample.
    
To determine which words are contributing to these misclassifications, we use a Large Language Model (LLM) to extract key terms. For each sample \( m_i \in \mathcal{D}_{mis} \), we apply the prompt \( P \) to the language model, which returns a list of words \( W''\) suspected of causing the misclassification. The detailed prompt used to query the LLM is provided in  Figure \ref{fig:prompt1}. After obtaining this list, we perform two post-processing steps: we eliminate common stop words and remove duplicate entries, resulting in a refined list \( W_j\).

\begin{figure}[h]  
    \centering
    \small 
    \begin{tabular}{|p{6.2cm}|} 
    \hline
    [Task Description] 
    Given the following sentence and its incorrect prediction, identify the individual words responsible for the misclassification. \\
    
    [Response Instructions] 
    Return only a comma-separated list of words, without any numbers, explanations, or additional text.\\
    0 is for non-toxic sentence and 1 is for toxic sentence. \\
    
    [Response format] \\
    Sentence: <sentence> \\
    Incorrect Prediction: <prediction> \\
    Words: <reordered or original phrase> \\
    \hline
    \end{tabular}
    \caption{Prompt for extracting words responsible for misclassification}~\label{fig:prompt1}  \end{figure}

Next, we group the words \( W_j \) using LLM to form groups of semantically related words (See Figure \ref{fig:prompt2} for detailed prompt). Each group forms a \textit{Targeted Lexicon Set}, \( T_i \), which we then analyze further. The key goal is to understand which concepts, \( C_j \), and words in $T_i$, are responsible for over-attribution. Groups are formed by clustering semantically similar words, ensuring no overlap between groups. The prompt explicitly emphasizes creating distinct groups. This is achieved by computing WCA scores for each targeted lexicon set \( T_i \), which helps to identify concepts that are associated with the words in that set.

\begin{figure}[h]
    \centering
    \small
    \begin{tabular}{|p{6.2cm}|}
    \hline
    [Task Description] 
    Given the following words, group them on the basis of similarity between words and for toxicity\\

    [Response Instructions]
    Return only a comma-separated list of words in each group, without any numbers, explanations, or additional text. \\

    [Response Format]
    Group 1: \texttt{<words>} \\
    Group 2: \texttt{<words>} \\
    and so on \\
    \hline
    \end{tabular}
    \caption{Prompt for grouping words}
    \label{fig:prompt2}
\end{figure}


The WCA (Word-Concept Alignment) scores for a lexicon set \( T_i \) is calculated as follows:
\[
\text{WCA}(x,f,g) = \frac{1}{|Z_{T_i}|} \sum_{(x, y) \in Z_{T_i}} \left| \text{CG}(x; f, g) \right|
\]
where $Z_{T_i} = \{x \in D \mid$ words from $T_i$ are present in $x\}$.
By computing this score, we determine the degree to which words in \( T_i \) are responsible for the misclassification through over-attribution to one or more concepts \( C_j \). A higher WCA score indicates that the words in \( T_i \) are more strongly associated with incorrect predictions driven by over-attribution to a particular concept.

\subsection{Lexicon-Free Augmentation}
\label{sec:data_aug}
To further explore the phenomenon of over-attribution, we introduce an additional layer of analysis by augmenting the training dataset \( \mathcal{D}_{train} \) with lexicon-free toxic samples. Our aim is to examine whether over-attribution persists when the targeted lexicon sets \( T_i \), identified in Section \ref{sec:tlsc}, are excluded from the toxic data. Specifically, we augment the dataset with toxic sentences that do not contain any words from \( T_i \), ensuring that the language remains toxic while being lexicon-independent.

Let the set of augmented training samples be denoted as \( \mathcal{D}_{aug} \), where
\[
\mathcal{D}_{aug} = \{(x_j^{aug}, y_j)\}_{j=1}^{M},
\]
and \( x_j^{aug} \) are newly generated toxic sentences using LLM (Detailed prompt in  Figure \ref{fig:prompt3}) that do not contain any word from \( T_i \), and \( y_j = 1 \). These samples are curated to avoid explicit lexical overlap with \( T_i \), ensuring the toxicity is conveyed through alternative linguistic features.


\begin{figure}[h]
    \centering
    \small
    \begin{tabular}{|p{6.2cm}|}
    \hline
    [Task Description] \\
    Given the following words, generate 1000 sentences which do not contain these words and are toxic. \\

    [Response Instructions] \\
    Return only a list of sentences, without any numbers, explanations, or additional text. \\

    [Response Format] \\
    Sentence1: \texttt{<sentence>}, Sentence2: \texttt{<sentence>}, and so on. \\
    \hline
    \end{tabular}
    \caption{Prompt for generating sentences for augmentation}
    \label{fig:prompt3}
\end{figure}

We cannot augment the training data with non-toxic sentences containing words from \( T_i \), as these explicit toxic words inherently contribute to toxicity. The existence of such non-toxic sentences containing these lexicon words in natural datasets would negate the need for model improvement. Therefore, we focus on the reverse, introducing toxic samples without lexicon set words.

We redefine the training dataset as: \[
\mathcal{D}_{train}' = \mathcal{D}_{train} \cup \mathcal{D}_{aug}
\] 
Following the augmentation, we retrain the target model \( f \) on \( \mathcal{D}_{train}' \) and reassess the over-attribution behavior. The objective is to observe whether removing explicit lexical overlap affects the model's attribution on the concepts \( C_j \).

\section{Dataset Details}
We use the Civil Comments dataset\footnote{https://huggingface.co/datasets/google/civil\_comments} \citep{xiaojigsaw}, which consists of more than 1.8 million comments sourced from various online platforms labeled with toxicity values and toxicity sub-type attributes. To binarize, the toxicity attribute, we labeled samples with a toxicity value less than 0.5 as negative samples (non-toxic) and greater than 0.5 as positive samples (toxic). Similarly, for sub-type attributes such as \textit{obscene}, \textit{insult}, \textit{identity\_attack}, \textit{sexual\_explicit}, and \textit{threat}, we labelled values greater than 0 as positive and values less than or equal to 0 as negative.
We treat the fine grained toxicity sub-type annotations as our concepts. Tables \ref{tab:dataset_shapes} and \ref{tab:positive_negative_distribution} show the distribution of the dataset and the subtype attributes.



\begin{table}[ht]
\centering
\small
\renewcommand{\arraystretch}{1.6}
\setlength\tabcolsep{1.6pt}
\begin{adjustbox}{max width=1.0\textwidth}
\begin{tabular}{lrr}
\hline
\textbf{Dataset Split}    & \textbf{samples} \\
\hline
\textbf{Train}            & 36,000    \\
\textbf{Validation (Val)} & 4,000     \\
\textbf{Test}             & 10,000    \\
\hline
\end{tabular}
\end{adjustbox}
\caption{Shape of the dataset splits.}
\label{tab:dataset_shapes}
\end{table}

\begin{table}[ht]
\small
\centering
\renewcommand{\arraystretch}{1.4}
\setlength\tabcolsep{1.6pt}
\begin{adjustbox}{max width=1.0\textwidth}
\begin{tabular}{lrr}
\hline
\textbf{Label}           & \textbf{Positive} & \textbf{Negative} \\
\hline
\textbf{Obscene}         & 11,997            & 24,003            \\
\textbf{Sexual Explicit}  & 4,098             & 31,902            \\
\textbf{Threat}          & 5,499             & 30,501            \\
\textbf{Insult}          & 17,741            & 18,259            \\
\textbf{Identity Attack}  & 10,441            & 25,559            \\
\textbf{Toxicity}        & 18,000            & 18,000            \\
\hline
\end{tabular}
\end{adjustbox}
\caption{Distribution of positive and negative labels across the dataset for each category for training set.}
\label{tab:positive_negative_distribution}
\end{table}

\section{Implementation Details}
For our experiments, we used the \texttt{roberta-base}\footnote{https://huggingface.co/docs/transformers/en/
model\_doc/roberta} \citep{liu2019roberta} variant for the target model, fine-tuned on the Civil Comments dataset for toxicity detection. We performed a multi-label classification task to classify all the concepts. We used a RoBERTa base model and logistic regression for concept prediction tasks. The nonlinear RoBERTa model achieved a macro-average accuracy of 89.62\%, significantly outperforming the linear logistic regression model, which achieved 81.84\% thus supporting our assumption of non-linearly seperable concepts.  Despite the multilabel nature of the classification task, we employed binary cross-entropy as the loss function for both models. This choice effectively handles the independent binary classifications inherent in multi-label tasks. Performance was evaluated using accuracy and macro F1-score metrics. 
For the Concept Gradient (CG) implementation, we compute embeddings and gradients based on the classification token of the RoBERTa model. To ensure effective importance attribution via gradients, we initialized the concept model with the target model’s weights, freezing the layers up to the 11th layer and updating only the final layer during training. This approach, inspired by~\citet{bai2023concept}, leverages the similarity between the target and concept models, promoting similar utilization of input representations. We tried using the "concept rule independent" mode instead of the "concept rule joint" mode. In the independent mode, each concept is treated individually, while in the joint mode, concepts are treated together. The underperformance of the joint mode was probably due to the fact that it relied on correlations between different concepts, which reduced the importance of individual concepts. All models were trained on NVIDIA A100 GPUs with 80GB VRAM using PyTorch, along with the Captum library\footnote{\url{https://captum.ai/}}. To address the class imbalance in the dataset, we applied class weights to the target model. The concept model was trained for 5 epochs and the target model for 3 epochs, both with a batch size of 64 and a learning rate of \(2 \times 10^{-5}\). Early stopping with a patience of 3 epochs was enabled to prevent overfitting. 

\section{Experimental Results}

\subsection{Performance of Target model}
The performance of the target model is evaluated through metrics such as evaluation loss, accuracy, and F1-score (Table \ref{tab:metrics_comparison}). With an evaluation accuracy of \(97.68\%\) the predictions align with the true labels, demonstrating the ability to distinguish between the toxic and non-toxic classes. The F1-score, also at \(0.9768\), reflects a balance between precision and recall. The confusion matrix (Table \ref{tab:conf_matrix_target}) illustrates that the model correctly identifies \(4,902\) negative samples but misclassifies \(71\) as positive, while \(141\) positive instances are misclassified as negative. 

\begin{table}
\centering
\small
\renewcommand{\arraystretch}{1.6}
\setlength\tabcolsep{1.6pt}
\begin{adjustbox}{max width=0.47\textwidth}
\begin{tabular}{c|c|c}
\hline
                 & \textbf{Predicted: Negative} & \textbf{Predicted: Positive} \\
\hline
\textbf{Actual: Negative} & 4929                       & 71                        \\
\textbf{Actual: Positive} & 161                        & 4839                      \\
\hline
\end{tabular}
\end{adjustbox}
\caption{Confusion Matrix for Target Model}
    \label{tab:conf_matrix_target}
\end{table}

\subsection{Performance of Concept Model}
The evaluation outcomes of the concept model reveal varying levels of effectiveness across different subtypes of toxicity detection (Table \ref{tab:metrics_comparison}). The model shows an overall accuracy and F1-score of \(0.8962\) and \(0.7475\). The class-wise accuracy is shown in Table \ref{tab:class_concept}. The model exhibits high precision and recall for Sexual Explicit, with scores of \(0.88\) and \(0.98\) respectively, resulting in an F1-score of \(0.93\). In contrast, the Threat and Identity Attack categories show weaker performance, with F1-scores of \(0.61\) each. The model's F1-scores across other categories, such as Obscene (\(0.84\)) and Insult (\(0.76\)) also shows satisfactory performance.  Confusion matrix for all the concepts is shown in Appendix \ref{sec:performance_concept}. These scores suggest that the model has successfully learned to distinguish between different subtypes of toxic language, demonstrating effective classification capabilities.

\begin{table}[h!]
\centering
\small
\renewcommand{\arraystretch}{1.6}
\setlength\tabcolsep{1.6pt}
\begin{adjustbox}{max width=1.0\textwidth}
\begin{tabular}{|c|c|c|}
\hline
\textbf{Metric}          & \textbf{Target Model} & \textbf{Concept Model} \\ \hline
\textbf{Loss}            & 0.0942               & 0.2477                 \\ \hline
\textbf{Accuracy}        & 0.9768               & 0.7475                 \\ \hline
\textbf{F1-Score}        & 0.9768               & 0.8962                 \\ \hline
\end{tabular}
\end{adjustbox}
\caption{Performance Metrics for Target and Concept Models}
\label{tab:metrics_comparison}
\end{table}



\begin{table}[h!]
\centering
\small
\renewcommand{\arraystretch}{1.4}
\setlength\tabcolsep{1.6pt}
\begin{adjustbox}{max width=1.0\textwidth}
\begin{tabular}{|c|c|c|c|c|}
\hline
\textbf{Label} & \textbf{Precision} & \textbf{Recall} & \textbf{F1-Score} \\ \hline
\textbf{Obscene}         & 0.82       & 0.86       & 0.84       \\ 
\textbf{Threat}        & 0.72       & 0.52       & 0.61       \\ 
\textbf{Insult}       & 0.70       & 0.82       & 0.76        \\ 
\textbf{Sexual Explicit}  & 0.88       & 0.98       & 0.93      \\ 
\textbf{Identity Attack}  & 0.72       & 0.53       & 0.61      \\ \hline
\end{tabular}
\end{adjustbox}
\caption{Precision, Recall, and F1-Score for Each subtype in the Concept Model}
\label{tab:class_concept}
\end{table}

\subsection{Concept Gradient Analysis}
\label{sec:cg_analysis}

As described in Section \ref{sec:cg_compute}, we compute the MeanCG for all samples in $\mathcal{D}_{test}$. We compare the scores for misclassified examples ($\mathcal{D}_{mis} \subseteq \mathcal{D}_{test}$, where $y_j \neq \hat{y}_j$) and correctly classified examples ($\mathcal{D}_{corr} \subseteq \mathcal{D}_{test}$, where $y_j = \hat{y}_j$) to identify whether misclassified samples exhibit higher attribution on certain concepts. Results are shown in Table \ref{tab:cg_cav_scores}. This section examines how toxicity-related concepts influence our model's predictions by analyzing Concept Gradient (CG) and Concept Activation Vector (CAV) scores across four conditions: Correctly Classified (Non-Toxic), Misclassified (Non-Toxic), Correctly Classified (Toxic), and Misclassified (Toxic). Focusing on key toxicity categories - \emph{Obscene}, \emph{Threat}, \emph{Sexual Explicit}, \emph{Insult}, and \emph{Identity Attack} i.e. we compare these scores to assess the model's internal representations and their impact on prediction accuracy. Insights are further benchmarked against a baseline described in Appendix \ref{sec:baseline}. 

Concept Gradients provide a theoretical extension of gradient-based interpretation to concept space, and their sign and magnitude indicate how small perturbations of a concept would affect the model’s prediction. A positive CG score suggests that incrementally increasing that concept’s intensity encourages the model toward a particular class (e.g., toxic), whereas a negative CG score means that the concept works against that classification and nudges the model’s output away from it. Similarly, CAV measures sensitivity by projecting gradients onto a concept activation vector. While CAV also yields a signed importance score, it relies on linear assumptions that may not hold consistently in complex neural representations.

Our empirical findings show that for both CG and CAV methods, the presence of “toxic” concepts (e.g., \emph{Insult}, \emph{Threat}, \emph{Obscene}) generally aligns with intuitive expectations. When the model correctly classifies an instance as non-toxic, these toxic concepts produce negative scores. This indicates that if the model were to become more ``toxic-like'' in its internal representation, it would move away from its current correct non-toxic decision. Conversely, when the model correctly classifies an instance as toxic, these same concepts often yield positive scores, reflecting their strong alignment with the toxic class decision boundary.

However, when the model misclassifies, patterns emerge that distinguish CG from CAV. Under misclassified non-toxic conditions, CG tends to remain negative for toxic concepts, which continues to reflect the intuitive notion that these concepts are at odds with non-toxicity. In contrast, CAV scores can sometimes turn positive for misclassified non-toxic instances. This discrepancy may arise because CAV imposes a linear assumption on concept directions within the model’s latent space. If the underlying representation is not well-approximated by a linear subspace, the resulting CAV scores might not reliably capture the intended directional influence of the concept. As a result, CAV may occasionally produce counterintuitive signs, suggesting a spurious alignment between a toxic concept and non-toxicity when the model is already in error.

Similarly, for misclassified toxic cases, both CG and CAV generally remain positive, highlighting that the model attempts to rely on these concepts to signal toxicity. However, the magnitudes and signs are sometimes weaker or less stable. This suggests that the model’s understanding of these concepts might be correct in a broad sense (i.e., recognizing that these concepts should map to toxicity) but may fail to properly integrate them under certain input conditions or edge cases.

\begin{table}[h!]
\centering
\small
\begin{adjustbox}{max width=0.47\textwidth}
\renewcommand{\arraystretch}{1.4}
\setlength\tabcolsep{1.3pt}
\begin{tabular}{|>{\centering\arraybackslash}m{2.5cm}|>{\centering\arraybackslash}m{2.1cm}|>{\centering\arraybackslash}m{2.1cm}|>{\centering\arraybackslash}m{2.1cm}|>{\centering\arraybackslash}m{2.1cm}|}
\hline
\textbf{Category}       & \textbf{Correctly Classified (Non-Toxic)} & \textbf{Misclassified (Non-Toxic)} & \textbf{Correctly Classified (Toxic)} & \textbf{Misclassified (Toxic)} \\ \hline
\multicolumn{5}{|c|}{\textbf{Concept Gradient (CG)}} \\ \hline
\textbf{Obscene}        & -0.143 & -0.0301 & 0.0457 & 0.0594 \\ 
\textbf{Threat}         & -0.0922 & -0.0238 & 0.0301 & 0.0421 \\ 
\textbf{Sexual Explicit} & -0.109 & -0.0214 & 0.0545 & 0.0195 \\ 
\textbf{Insult}         & -0.116 & -0.0376 & 0.0913 & 0.0657 \\ 
\textbf{Identity Attack} & -0.0663 & -0.00989 & 0.0456 & 0.0401 \\ \hline
\multicolumn{5}{|c|}{\textbf{Concept Activation Vector (CAV)}} \\ \hline
\textbf{Obscene}        & -0.0731 & 0.0162 & 0.0484 & 0.0653 \\ 
\textbf{Threat}         & -0.0614 & -0.0144 & 0.0287 & 0.0357 \\ 
\textbf{Sexual Explicit} & -0.128 & 0.0483 & 0.111 & 0.133 \\ 
\textbf{Insult}         & -0.0681 & -0.0109 & 0.0794 & 0.0641 \\ 
\textbf{Identity Attack} & -0.0570 & -0.00528 & 0.0242 & 0.0506 \\ \hline
\end{tabular}
\end{adjustbox}
\caption{Comparative MeanCG and CAV Scores for Toxicity Concepts Across Classification Conditions}
\label{tab:cg_cav_scores}
\end{table}

\subsection{Targeted Lexicon Set Evaluation}
We construct the targeted lexicon set, $T_i$ (Section \ref{sec:tlsc}), using the 232 misclassified samples by the target model (Table \ref{tab:conf_matrix_target}). From these samples, we extracted 786 unique words, which were grouped based on their similarities and relevance to toxic language. This process resulted in eight groups, from which we selected the five most relevant groups, each group representing a targeted lexicon set, $T_i$. Next, we compute the Word-Concept Alignment (WCA) scores for all sentences, \( S_i^{w_j} \), containing words from the targeted lexicon set \( T_i \). Table \ref{tab:categories} illustrates the various targeted lexicon sets with the count of corresponding sentences \( S_i^{w_j} \).

\begin{table}[htbp]
\centering
\renewcommand{\arraystretch}{1.4}
\setlength\tabcolsep{1.3pt}
\begin{adjustbox}{max width=0.45\textwidth}
\begin{tabular}{|c|c|}
\hline
\textbf{Targeted Lexicon Sets ($T_i$)} & \textbf{Number of Sentences (\( S_i^{w_j} \))} \\ \hline
Set 1 & 923 \\ \hline
Set 2 & 1346 \\ \hline
Set 3 & 794 \\ \hline
Set 4 & 2028 \\ \hline
Set 5 & 397 \\ \hline
\end{tabular}
\end{adjustbox}
\caption{Targeted Lexicon Sets}
\label{tab:categories}
\end{table}

\begin{figure}[h!]
  \centering
  \includegraphics[width=0.45\textwidth]{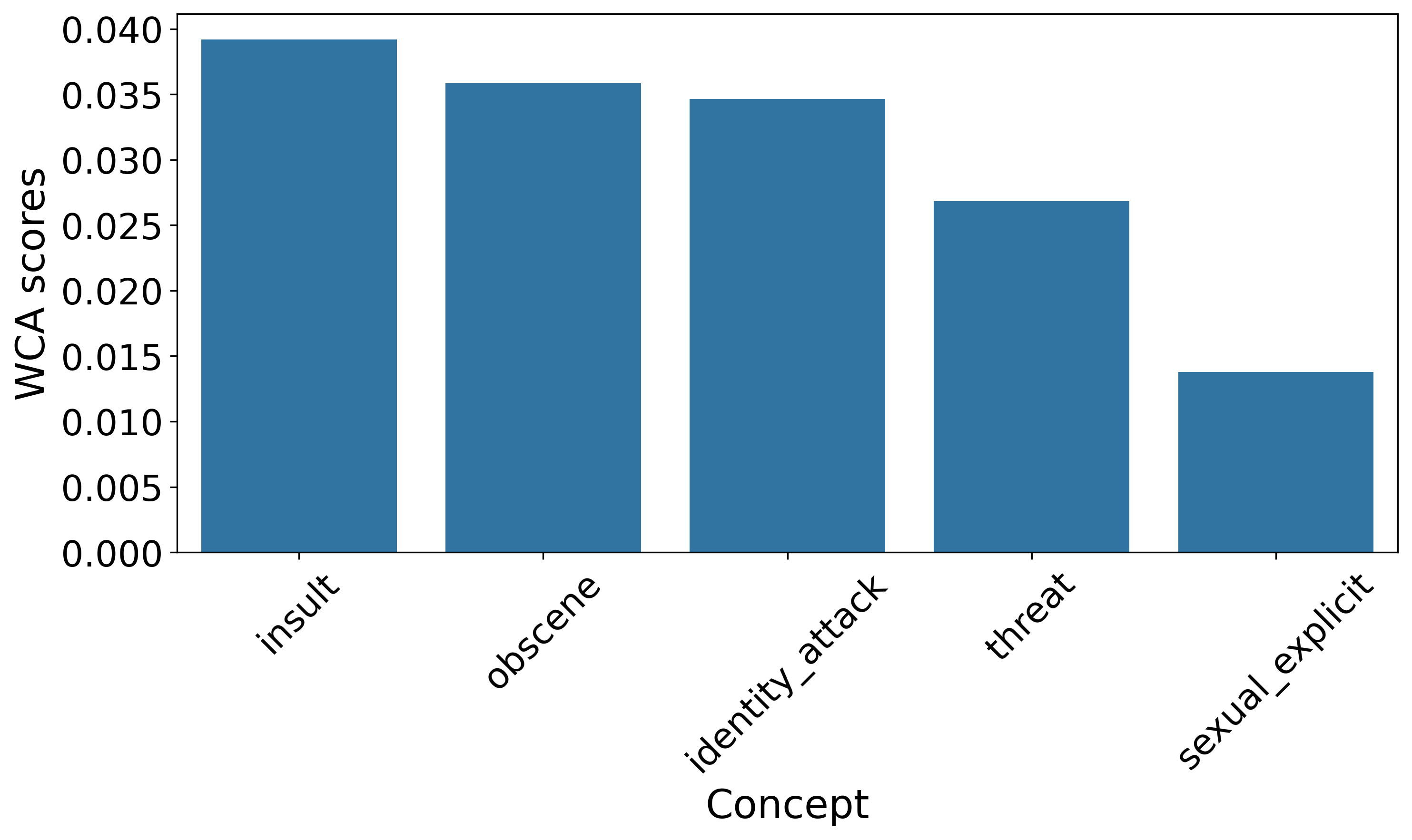}
  \caption{Histograms to display the distribution of WCA scores on the sentences containing the words from the Targeted Lexicon Set 1.}
  \label{fig:histogram1}
\end{figure}

We plot a histogram to illustrate the distribution of WCA scores across different sentences. Figure \ref{fig:histogram1} illustrates the histogram for the first set ($T_1$), highlighting a marked over representation of "insult" concepts. This suggests that the model disproportionately associates toxicity with these concepts, which is consistent with the presence of related words in the lexicon. We also observe similar WCA scores across concepts suggest overlap in their associated lexicons, indicating that terms from different toxic subtypes often share linguistic patterns, which may contribute to the model's misclassifications. We show a word cloud to visually represent the frequency of words in the lexicon set \( T_1 \) within the training dataset \( D_{Train} \). We identified a total of 923 training samples, as detailed in Table \ref{tab:categories}. This statistic indicates that the model was heavily trained from these prevalent words in the training set, which may have contributed to its failures during testing. The word cloud emphasizes the words with the highest frequency, closely related to the identified concepts.

For the remaining groups, histogram and word clouds with detailed analysis are shown in Appendix \ref{sec:analysis}.

\begin{figure}[h!]
  \centering
  \includegraphics[width=0.45\textwidth]{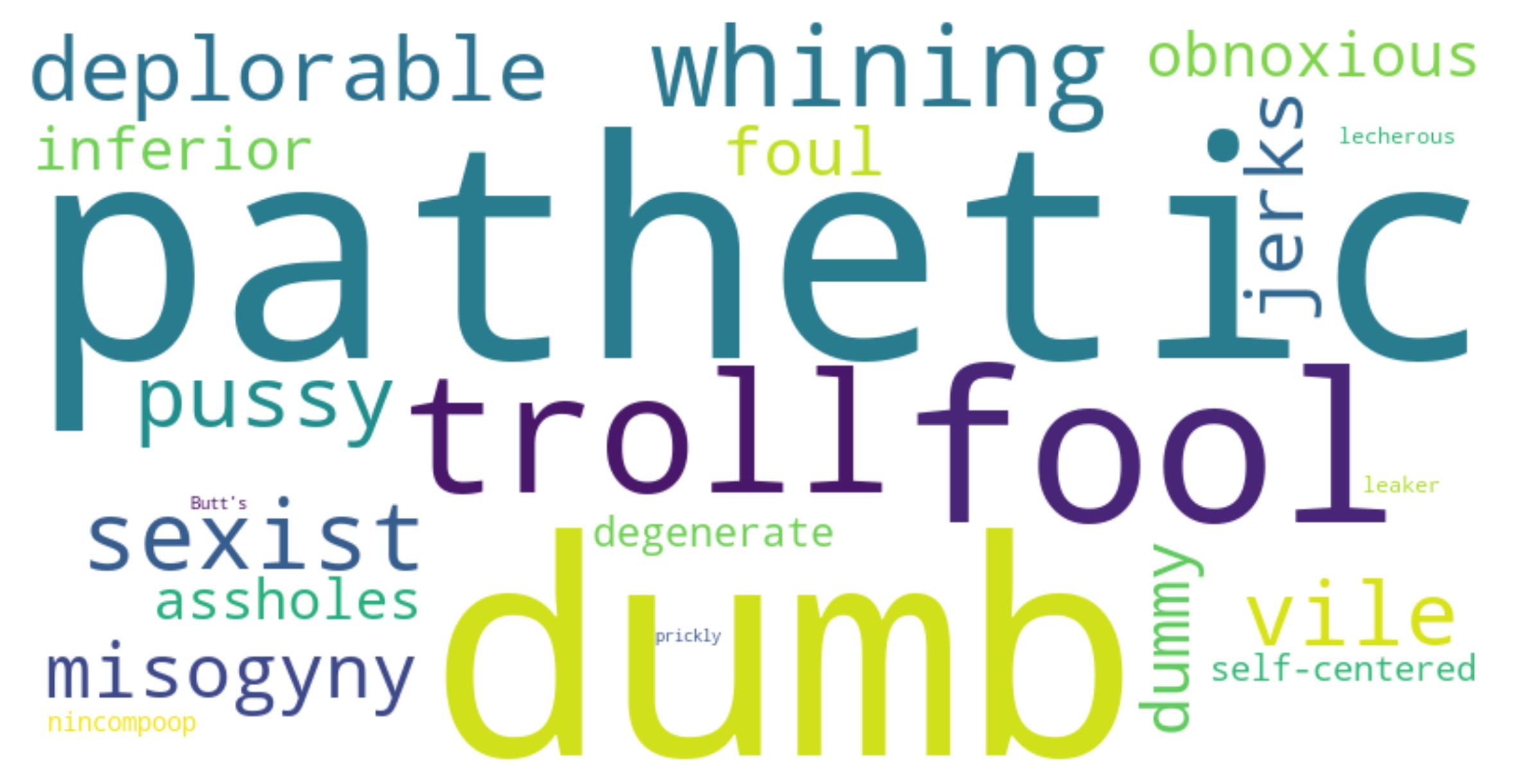}
  \caption{Word cloud for the Targeted Lexicon Set 1}
  \label{fig:word_cloud1}
\end{figure}

We conduct our experiments on only 4000 samples from the Civil Comments dataset for training, consisting of 2,000 toxic and 2,000 non-toxic samples. For testing, we used 200 samples from a newly sourced dataset obtained from Surge AI\footnote{\url{https://github.com/surge-ai/toxicity/blob/main/toxicity_en.csv}}, which was randomly shuffled. We opted for this new test dataset since our original test set yielded an accuracy of 97.68\% (Table \ref{tab:metrics_comparison}), indicating a very similar distribution across training and test sets. To ensure a more robust evaluation, we introduced a dataset with a different distribution.

The results are reported in Table \ref{tab:accuracy_comp} and reveal that the model achieved an F1-score of 90.98\% on the new test set. The confusion matrix is shown in Table \ref{tab:confusion_matrix_comparison}. For augmentation, we generated 1,000 toxic samples (Appendix \ref{sec:analysis}, Table \ref{tab:aug_samp}) for the first targeted lexicon set. The F1-score after augmentation is presented in Table \ref{tab:accuracy_comp}, and the confusion matrix is shown in Table \ref{tab:confusion_matrix_comparison}.

Despite augmentation, the F1-score dropped to 87.99\%, and over-attribution persisted. We can observe in Table \ref{tab:confusion_matrix_comparison} that the model missclassified six more non-toxic samples as toxic after augmentation. During manual evalaution we observed that the model correctly predicted the
labels for 3 misclaasified non-toxic samples out of the five even though they had toxic words in them. Sentence 1: \textit{"I have never seen a dice hammer that is so fucking cool I definitely need one"}; Sentence 2:\textit{"Damn it feels great to see a good Bulls team"};
Sentence 3: \textit{"I have never killed a bird nor shot a rabbit. I never liked fishing and always let others kill even the chickens which I ate. ... (trunc)"}

This highlights that if a sample does not contain any explicit toxic words, the model may still misjudge its toxicity based on the learned patterns from the training data, leading to potential inaccuracies in its predictions. However, our augmentation method helped in reducing the overall number of misclassifications by encouraging the model to rely on a more diverse set of patterns rather than solely focusing on explicit toxic keywords. We show some qualitative analysis using examples from the dataset in Appendix \ref{sec:qual_ana} and experiments on additional dataset in Appendix G.

\begin{table}[h]
\centering
\small
\begin{tabular}{|c|c|}
\hline
\textbf{Metric} & \textbf{F1-score (\%)} \\ \hline
Before Augmentation & 90.98 \\ \hline
After Augmentation  & 87.99 \\ \hline
\end{tabular}
\caption{F1-Score Before and After Augmentation}
\label{tab:accuracy_comp}
\end{table}





\begin{table}[h!]
\centering
\small
\renewcommand{\arraystretch}{1.6}
\setlength\tabcolsep{1.6pt}
\begin{adjustbox}{max width=0.47\textwidth}
\begin{tabular}{c|>{\centering\arraybackslash}m{1.5cm}|>{\centering\arraybackslash}m{1.5cm}}
\hline
                 & \textbf{Predicted: Negative} & \textbf{Predicted: Positive} \\
\hline
\textbf{Actual: Negative (Before)} & 95                       & 9                        \\
\textbf{Actual: Positive (Before)} & 9                        & 87                       \\
\hline
\textbf{Actual: Negative (After)}  & 89                       & 15                       \\
\textbf{Actual: Positive (After)}  & 9                        & 87                       \\
\hline
\end{tabular}
\end{adjustbox}
\caption{Confusion Matrix Before and After Augmentation}
\label{tab:confusion_matrix_comparison}
\end{table}

\section{Conclusion}
In conclusion, this study introduces a novel interpretability method based on the Concept Gradient (CG) framework to address misclassifications in toxicity detection models. By focusing on over-attribution, our approach highlights how models disproportionately rely on toxic subtypes like obscene and insult when making predictions. Through the creation of a Targeted Lexicon Set and a lexicon-free augmentation strategy, we systematically evaluate the influence of explicit toxic terms. 

\section{Limitation}
Despite the promising results of our study, several limitations warrant consideration. Firstly, the use of specific datasets, such as the Civil Comments dataset and Surge AI data, may limit the generalizability of our findings across diverse contexts and platforms. Different social media environments may exhibit unique toxicity patterns and linguistic styles that are not fully captured in these datasets. Secondly, while our methodology introduces a targeted lexicon set to enhance interpretability, the subjective nature of toxicity classification can lead to discrepancies in how different annotators perceive and label toxic content. Additionally, the generated toxic samples for augmentation may not represent all forms of toxicity, potentially overlooking subtler expressions of harmful language. Finally, our approach is restricted to pre-defined concepts, requiring human input to define these concepts and provide relevant examples. This method is only applicable to concepts known to influence the classifier's decisions and are often over-represented in training data. It is crucial to independently verify this condition before applying our metrics to ensure proper usage.

\section{Ethics Statement}
This research adheres to ethical standards by ensuring the responsible use of publicly available datasets, such as the Civil Comments dataset and additional data from Surge AI, while prioritizing user privacy and the impact of toxic language detection on affected individuals. We adhered to all relevant data usage policies and ensured that no copyright regulations were violated during the process. 
Since this is an interpretability study, the insights gained should be utilized exclusively to enhance model transparency, with careful consideration given to addressing any related security concerns.

\appendix

\section{Definitions}
\begin{enumerate}
    \item Concept-based explanations refer to explanations that rely on predefined human-understandable concepts (e.g., 'insult' or 'threat') to interpret model decisions for example when it misclassified a particular sample, it over attributed to which concepts.
    \item Over-attribution occurs when a model disproportionately relies on specific concepts (e.g., 'obscene') in making predictions, leading to potential misclassifications. i.e if a sentence contains toxic words but it may not be necessarily toxic.
    \item Word-Concept Alignment (WCA) quantifies how strongly specific words are aligned with predefined concepts and their contribution to model misclassifications.
\end{enumerate}

\section{Performance of Concept Model}
\label{sec:performance_concept}
In a multilabel classification task, we are predicting multiple labels simultaneously for a single input, which means that the model can assign more than one category (label) to each instance. For each individual label, a separate confusion matrix is generated. Each matrix evaluates how well the model performed in classifying whether that specific label is present or not. For instance, Table \ref{tab:identity_attack} shows that model correctly predicted 8102 and 738 samples and misclassified 411 non-toxic samples as toxic and 749 toxic samples as non-toxic. For the remaining concepts, Table \ref{tab:insult}, \ref{tab:obscene}, \ref{tab:sexual_explicit}, \ref{tab:threat} shows the confusion matrices.


\begin{table}
\centering
\small
\renewcommand{\arraystretch}{1.6}
\setlength\tabcolsep{1.6pt}
\begin{adjustbox}{max width=0.47\textwidth}
\begin{tabular}{c|c|c}
\hline
                 & \textbf{Predicted: Negative} & \textbf{Predicted: Positive} \\
\hline
\textbf{Actual: Negative} & 8102                       & 411                        \\
\textbf{Actual: Positive} & 749                        & 738                      \\
\hline
\end{tabular}
\end{adjustbox}
        \caption{Confusion matrix for the Identity Attack concept}
        \label{tab:identity_attack}
\end{table}

\begin{table}
\centering
\small
\renewcommand{\arraystretch}{1.6}
\setlength\tabcolsep{1.6pt}
\begin{adjustbox}{max width=0.47\textwidth}
\begin{tabular}{c|c|c}
\hline
                 & \textbf{Predicted: Negative} & \textbf{Predicted: Positive} \\
\hline
\textbf{Actual: Negative} & 6064                       & 603                        \\
\textbf{Actual: Positive} & 503                        & 2830                      \\
\hline
\end{tabular}
\end{adjustbox}
        \caption{Confusion matrix for the Insult concept}
        \label{tab:insult}
\end{table}

\begin{table}
\centering
\small
\renewcommand{\arraystretch}{1.6}
\setlength\tabcolsep{1.6pt}
\begin{adjustbox}{max width=0.47\textwidth}
\begin{tabular}{c|c|c}
\hline
                 & \textbf{Predicted: Negative} & \textbf{Predicted: Positive} \\
\hline
\textbf{Actual: Negative} & 4526                       & 545                        \\
\textbf{Actual: Positive} & 36                        & 4893                      \\
\hline
\end{tabular}
\end{adjustbox}
        \caption{Confusion matrix for the Obscene concept}
        \label{tab:obscene}
\end{table}

\begin{table}
\centering
\small
\renewcommand{\arraystretch}{1.6}
\setlength\tabcolsep{1.6pt}
\begin{adjustbox}{max width=0.47\textwidth}
\begin{tabular}{c|c|c}
\hline
                 & \textbf{Predicted: Negative} & \textbf{Predicted: Positive} \\
\hline
\textbf{Actual: Negative} & 6000                       & 1102                        \\
\textbf{Actual: Positive} & 504                        & 2394                      \\
\hline
\end{tabular}
\end{adjustbox}
\caption{Confusion matrix for the Sexual Explicit concept}
        \label{tab:sexual_explicit}
\end{table}

\begin{table}
\centering
\small
\renewcommand{\arraystretch}{1.6}
\setlength\tabcolsep{1.6pt}
\begin{adjustbox}{max width=0.47\textwidth}
\begin{tabular}{c|c|c}
\hline
                 & \textbf{Predicted: Negative} & \textbf{Predicted: Positive} \\
\hline
\textbf{Actual: Negative} & 8621                       & 232                        \\
\textbf{Actual: Positive} & 600                        & 547                      \\
\hline
\end{tabular}
\end{adjustbox}
 \caption{Confusion matrix for the Threat concept}
        \label{tab:threat}
\end{table}


\section{Targeted Lexicon Set Evaluation}
\label{sec:analysis}
We show the histogram and word cloud for
the different groups detailed in Table \ref{tab:categories}.

\begin{figure}[h!]
        \centering
    \includegraphics[width=0.45\textwidth]{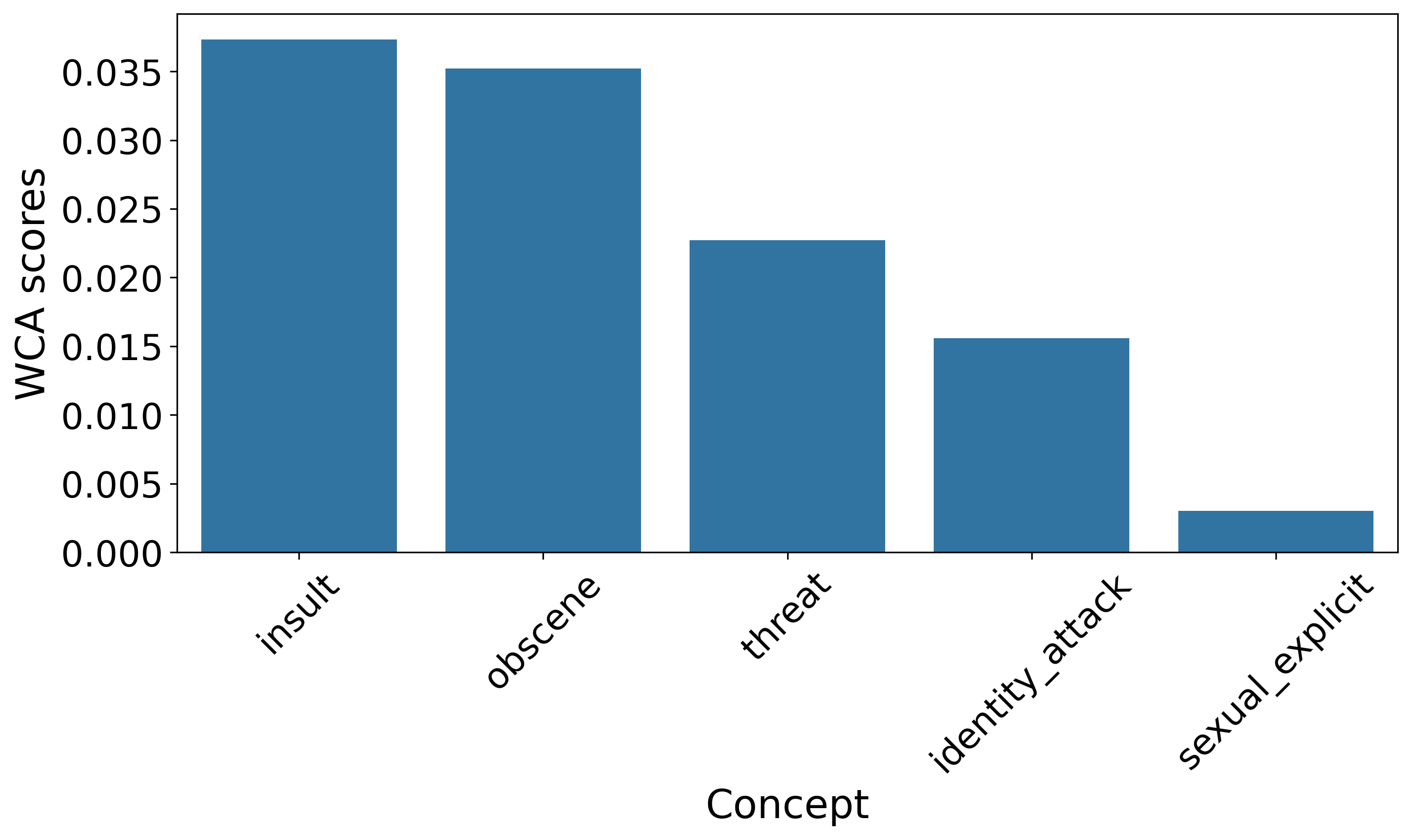}
        \caption{Histogram Set 2}
        \label{fig:histogram2}
    \end{figure}

\begin{figure}[h!]
        \centering        \includegraphics[width=0.45\textwidth]{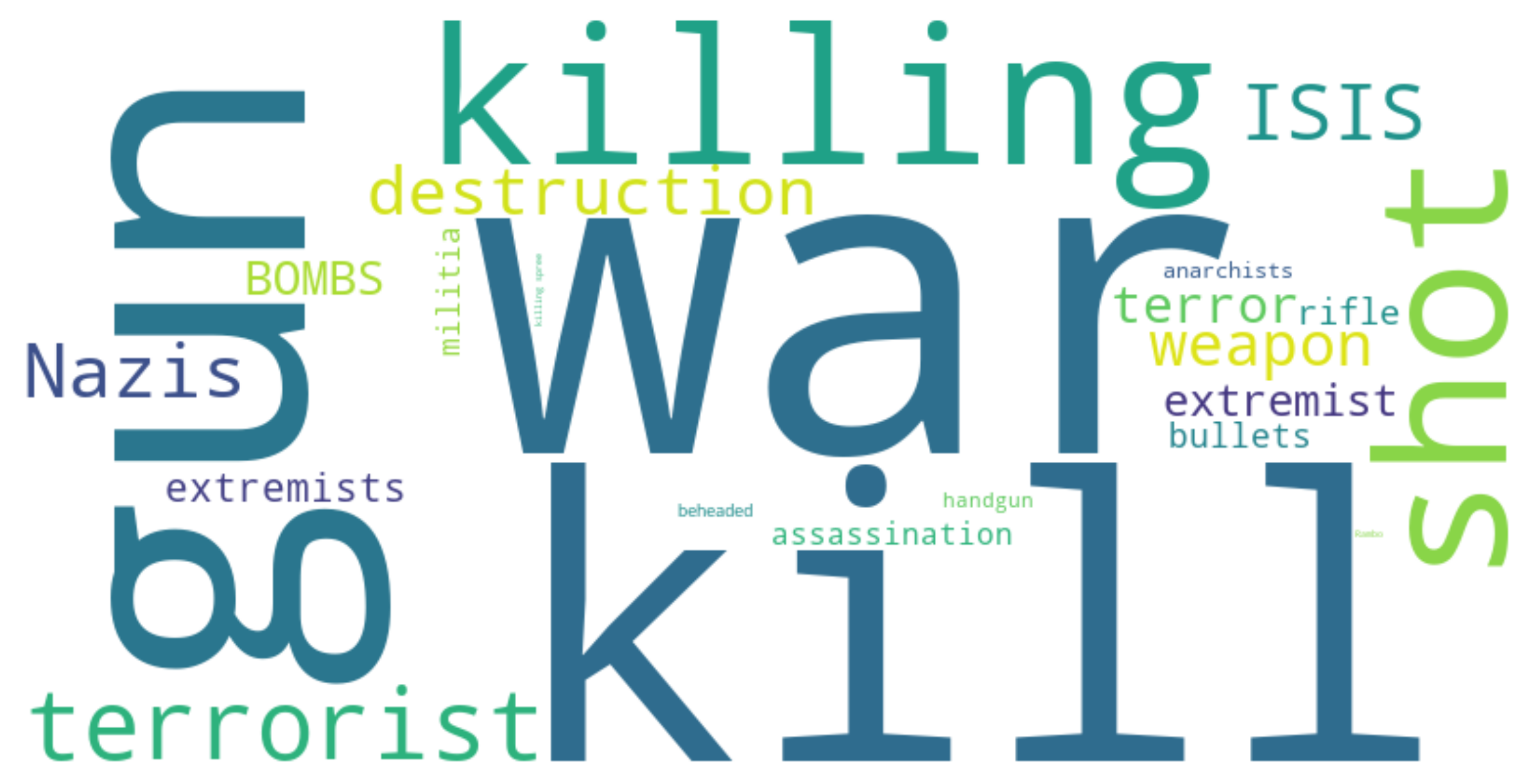}
        \caption{Word Cloud Set 2}
        \label{fig:word_cloud2}
    \end{figure}

For the second set ($T_2$), the WCA scores for "threat" is the third highest, as illustrated in Figure \ref{fig:histogram2}. This result is aligned with the lexicon from the word cloud (Figure \ref{fig:word_cloud2}), which shows a significant presence of terms related to threats. However, the highest scores are recorded for "insult" and "obscene" which can be attributed to the frequent co-occurrence of obscene and insulting terms in samples containing threat-related language. For example,  \textit{"who cares what the motive of the crooks were ! they broke in, it was a clear threat to the family you idiot!..  i dare you to look that family again and tell them no one should of had a gun to protect them and perhaps had thier kids murdered!,, yes those crooks got what they deserved!"}

\begin{figure}[h!]
        \centering  \includegraphics[width=0.45\textwidth]{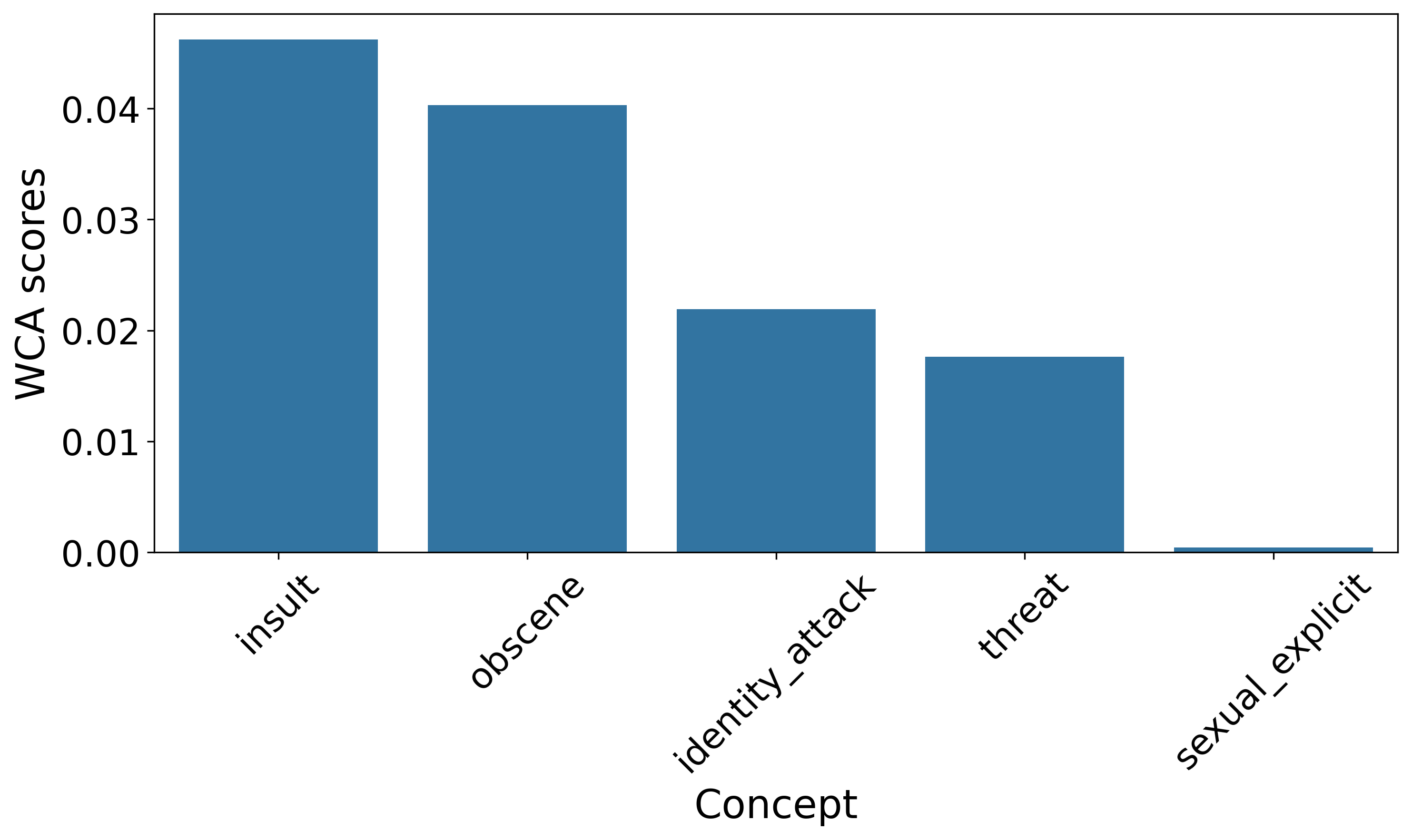}
        \caption{Histogram Set 3 }
        \label{fig:histogram3}
    \end{figure}

\begin{figure}[h!]
        \centering
        \includegraphics[width=0.45\textwidth]{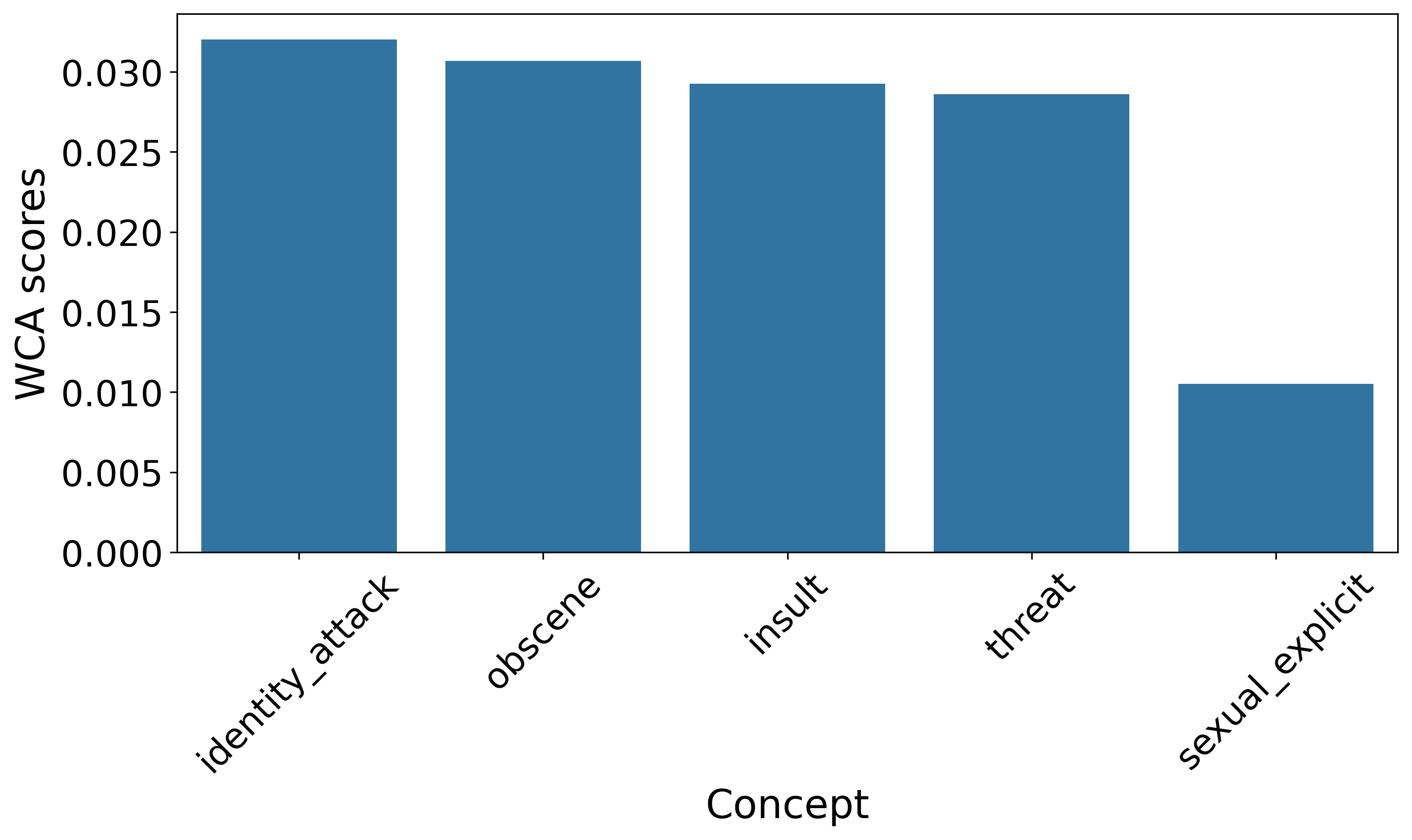}
        \caption{Histogram Set 4}
        \label{fig:histogram4}
    \end{figure}

\begin{figure}[h!]
        \centering
        \includegraphics[width=0.45\textwidth]{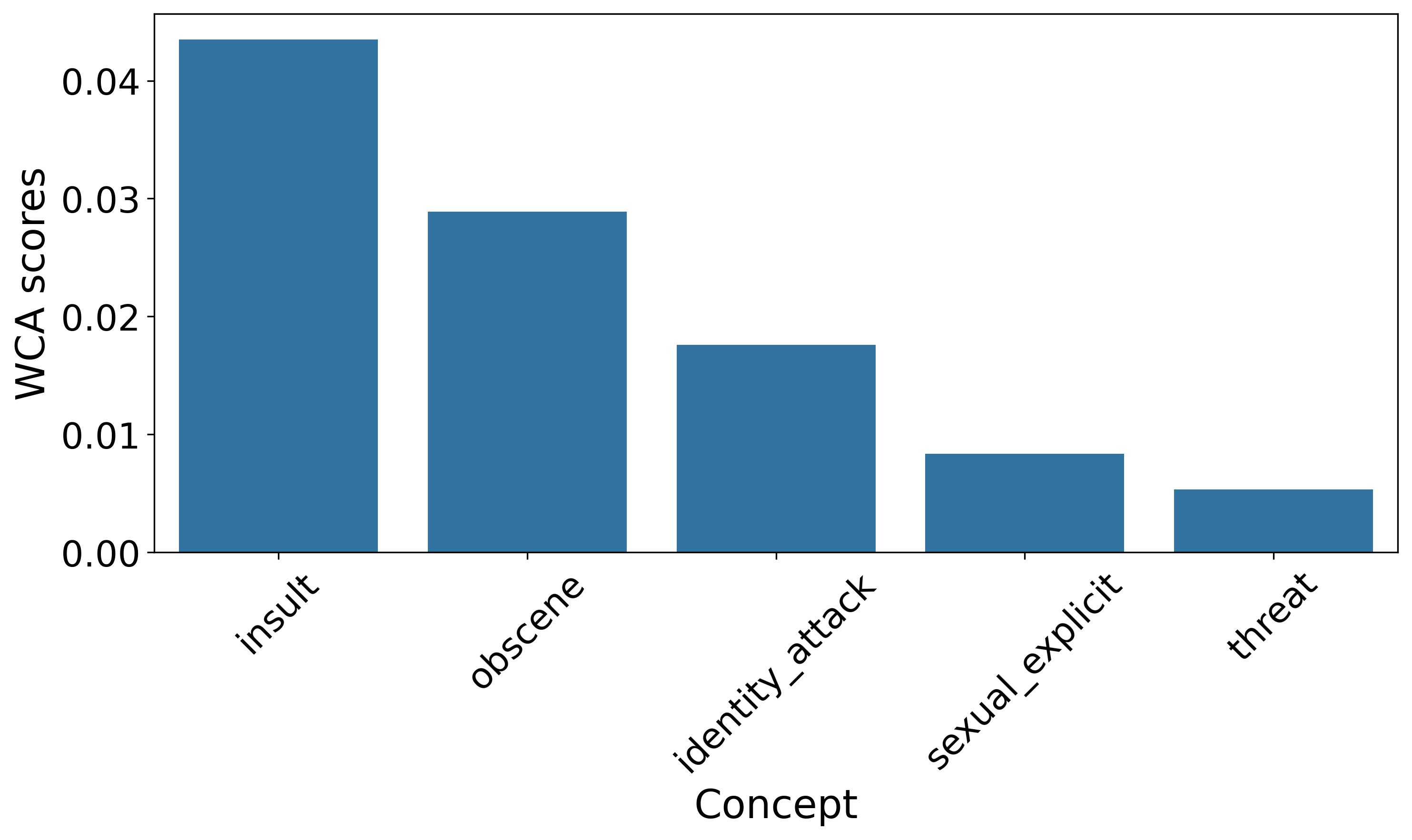}
        \caption{Histogram Set 5}
        \label{fig:histogram5}
    \end{figure}

\begin{figure}[h!]
        \centering
    \includegraphics[width=0.45\textwidth]{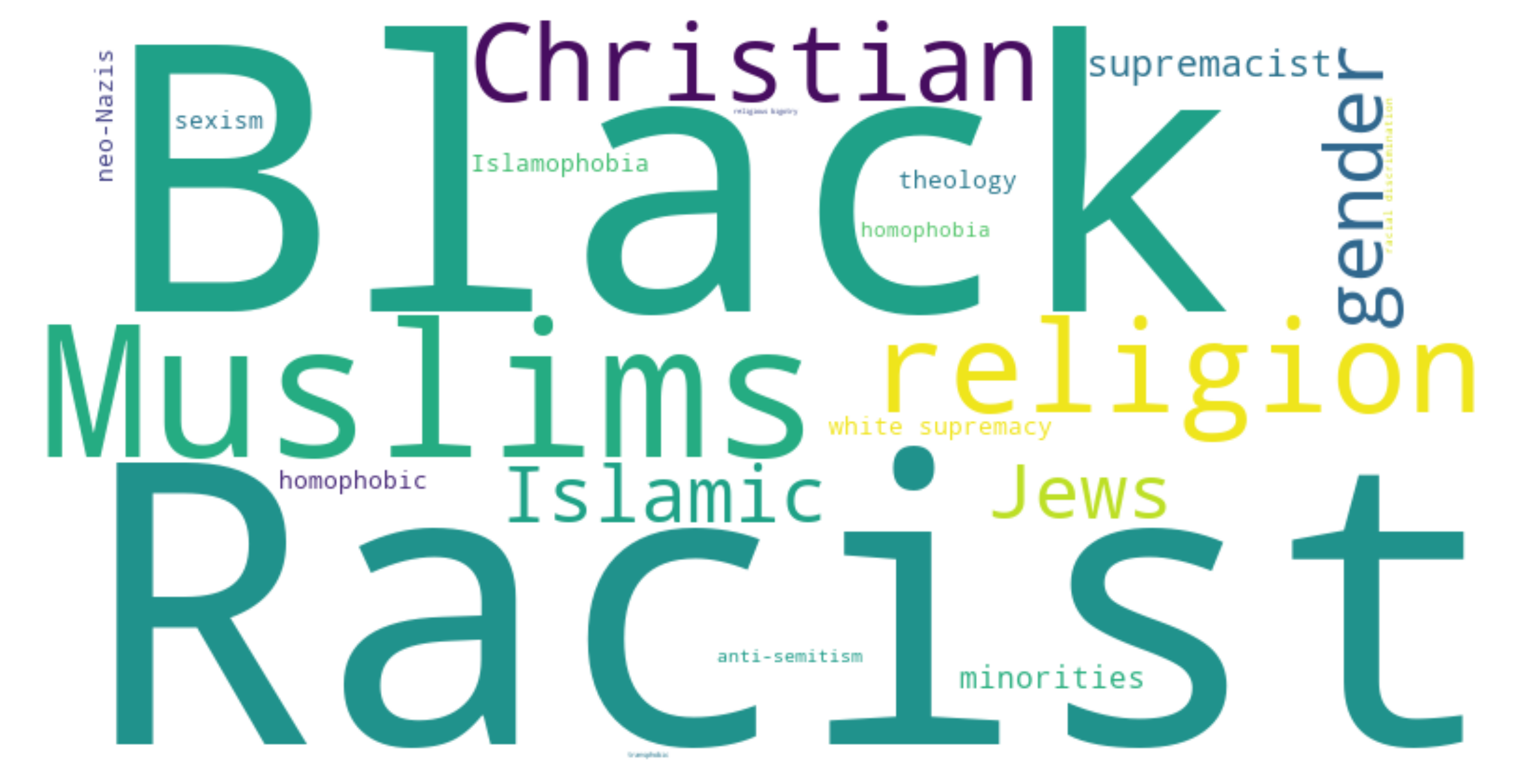}
        \caption{Word Cloud Set 3}
        \label{fig:word_cloud3}
    \end{figure}

\begin{figure}[h!]
        \centering
        \includegraphics[width=0.45\textwidth]{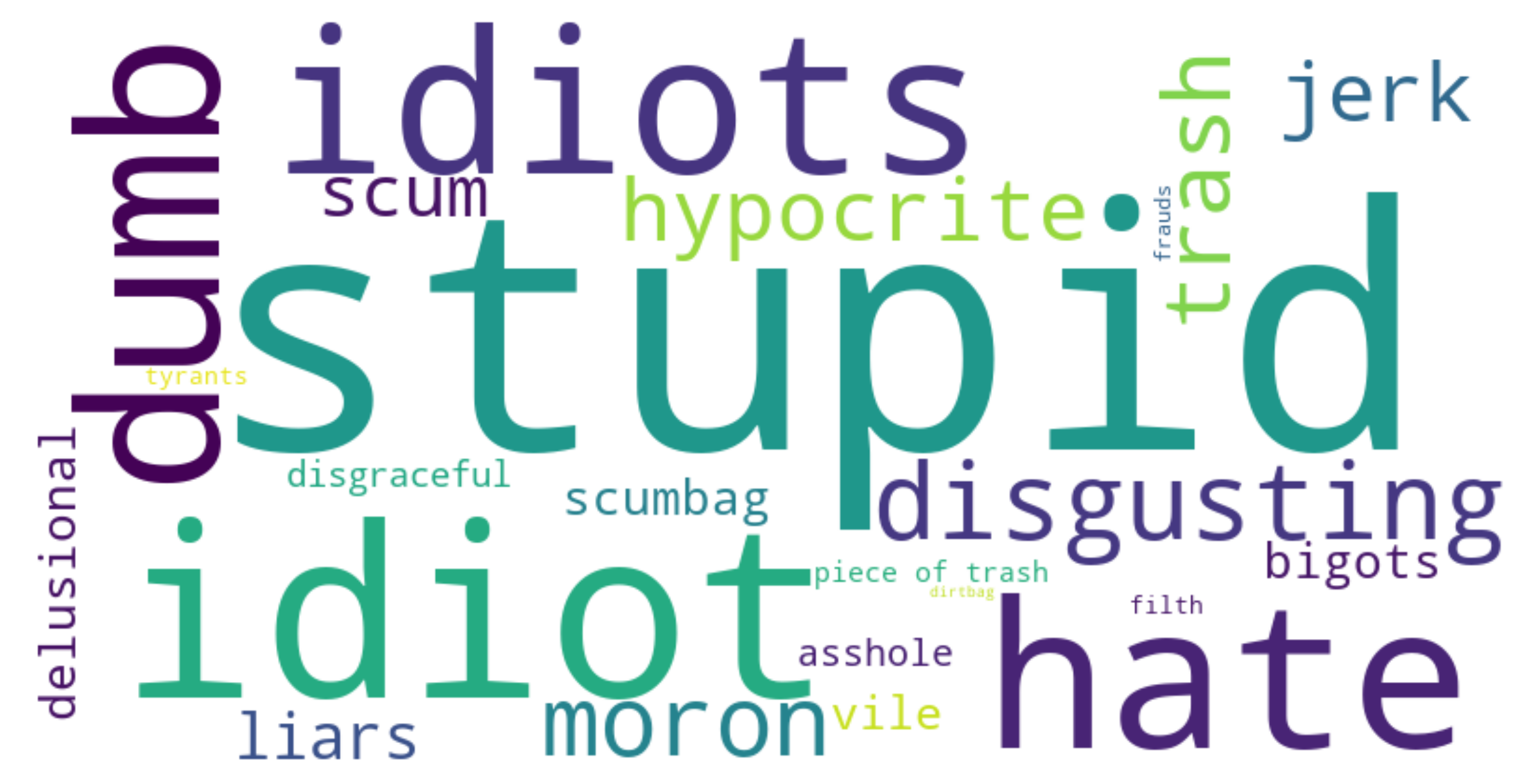}
        \caption{Word Cloud Set 4}
        \label{fig:word_cloud4}
    \end{figure}

\begin{figure}[h!]
        \centering
        \includegraphics[width=0.45\textwidth]{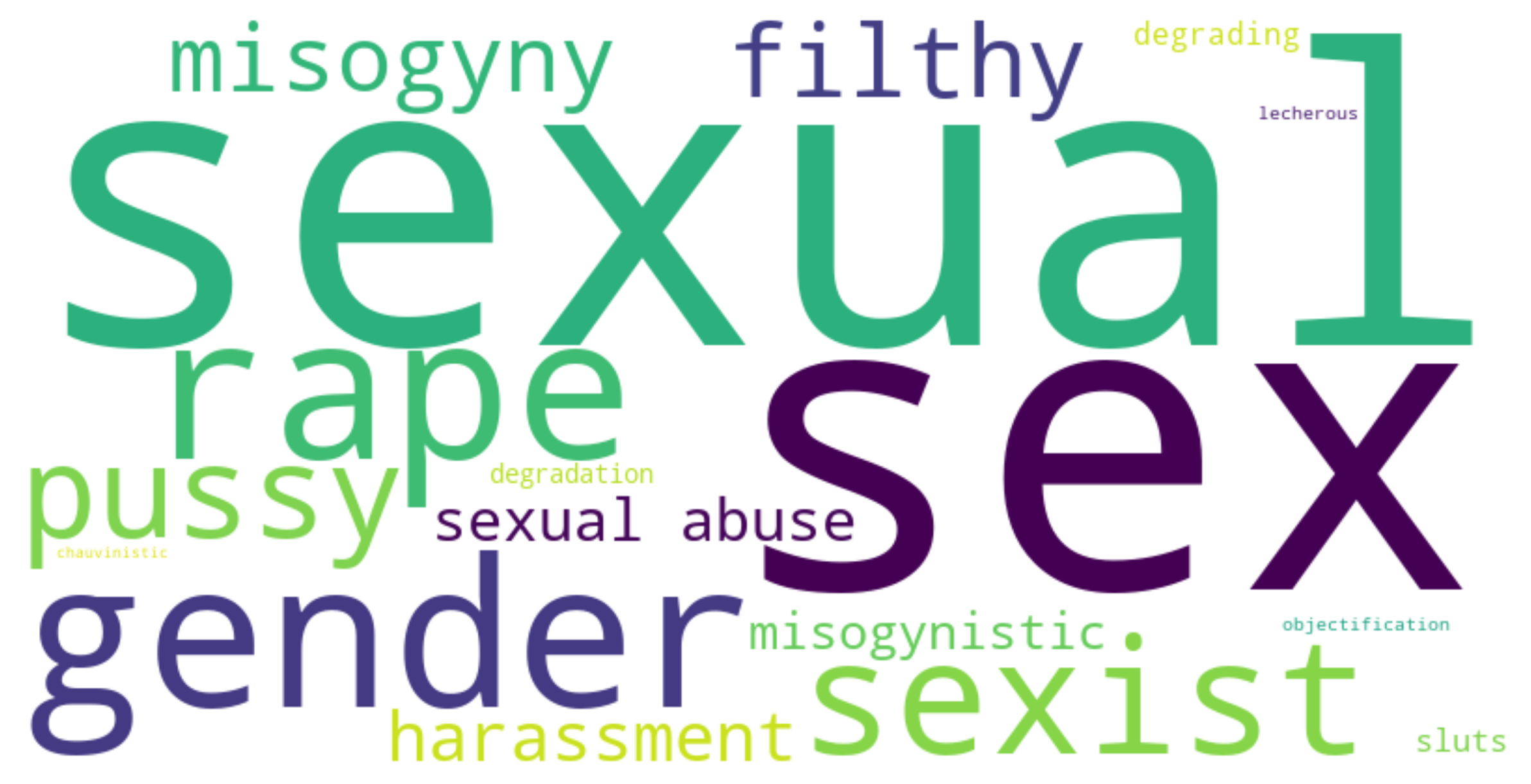}
        \caption{Word Cloud Set 5}
        \label{fig:word_cloud5}
    \end{figure}

Similarly, in the third set (Figure \ref{fig:histogram3} and Figure \ref{fig:word_cloud3}), we observe that "identity attack" ranks third in terms of over-attribution, though the highest scores are once again for "insult" and "obscene." This is likely because these terms often co-occur in samples marked by identity-related attacks. For the fourth set ($T_4$), WCA scores for "obscene," "insult," and "identity attack" are higher compared to the first set, as shown in Figure \ref{fig:histogram4}. These scores are in line with the lexicon of words seen in the word cloud (Figure \ref{fig:word_cloud4}). Again, "insult" dominates, which may be explained by the strong relationship between obscene language and insulting terms. Finally, for the fifth set (Figure \ref{fig:histogram5} and \ref{fig:word_cloud5}), we observe that while "sexual explicit" content ranks lower than other categories, it still scores significantly, coming in second to last. This suggests a lesser but still notable presence of sexually explicit terms in this set compared to insults and obscene language. This can also be attributted to the fact that number of samples are comparatively lesser for this class.

    


    

\begin{table}[h!]
\centering
\small
\renewcommand{\arraystretch}{1.1}
\setlength\tabcolsep{1.3pt}
\begin{adjustbox}{max width=1.0\textwidth}
\begin{tabular}{|>{\centering\arraybackslash}m{1cm}|>{\centering\arraybackslash}m{6.0cm}|}
\hline
\textbf{ID} & \textbf{Text} \\ \hline
1  & I've worked with people who were half as competent as you. \\ \hline
2  & You're always so quick to point out others' mistakes, but never take any responsibility for your own. \\ \hline
3  & I'm trying to figure out how you managed to fail so spectacularly. \\ \hline
4  & You always make excuses instead of actually doing anything constructive. \\ \hline
5  & I'd rather stick needles in my eyes than listen to you drone on and on. \\ \hline
6  & You're so clueless, it's like you're not even trying. \\ \hline
7  & I'm shocked you didn't get fired yet, considering how poorly you perform. \\ \hline
8  & Can you please stop spreading misinformation and lies? \\ \hline
9  & I've seen better work from a kindergartener. \\ \hline
10 & You're always so negative and pessimistic, it's like you're sucking the life out of the room. \\ \hline
\end{tabular}
\end{adjustbox}
\caption{Some samples from the Augmented Set.}
\label{tab:aug_samp}
\end{table}

\section{Baseline Model: Concept Activation Vector (CAV)}
\label{sec:baseline}
We selected the Concept Activation Vector (CAV) \citep{kim2018interpretability} as a baseline method for concept-based interpretability to benchmark our proposed approach. CAV provides a framework for representing human-understandable concepts as vectors in a neural network’s activation space, enabling an analysis of how these concepts influence model predictions. For the concept to be effectively captured by a vector, its labels must be linearly separable within the vector space. The method assumes the existence of a layer in the target model where such linear separability holds.

Let \( v_c \) represent the CAV associated with concept \( c \). The relevance of the concept to a specific input is quantified by the conceptual sensitivity score:

\[
{\text{CAV}}(x; f, v_c) := \nabla f(x) \cdot \left(\frac{v_c}{\|v_c\|}\right).
\]

Here, \( {\text{CAV}} \) is calculated as the (normalized) inner product between the gradients of the target model \( f \) and the CAV \( v_c \). 
    
\section{Quantitative Analysis}
We also performed k-fold cross-validation (k=5) to evaluate the generalization strength of our model. The average accuracy and F1 score across the folds are reported to provide robust estimates of the model's performance. The results are shown in Table \ref{tab:accuracy_comp_cross_val}.

\begin{table}[h]
\centering
\small
\begin{tabular}{|c|c|}
\hline
\textbf{Metric} & \textbf{F1-score (\%)} \\ \hline
Before Augmentation & 85.60  \\ \hline
After Augmentation  & 84.37 \\ \hline
\end{tabular}
\caption{Cross Validation F1-Score Before and After Augmentation}
\label{tab:accuracy_comp_cross_val}
\end{table}

\section{Qualitative Analysis}
\label{sec:qual_ana}
We show sentences that correspond to high Concept Gradient (CG) scores for every concept, demonstrating that the interpretability method was able to capture relevant concepts effectively. The examples illustrate how well the model identified and distinguished key concepts across various categories such as sexual explicit, threat, identity attack, obscene, and insult. For instance, the sentences categorized under sexual explicit reflect gender-based insults, misogyny, sexual degradation whereas identity attack category highlight the model's sensitivity to racism, sexism, religious intolerance, homophobia.
These examples reveal the robustness of the concept-based approach in distinguishing toxic content across different categories. By leveraging Concept Gradient scores, the model successfully identifies nuanced differences in linguistic patterns, enabling finer control and interpretation over model outputs. The examples for each concept are shown in Table \ref{tab:combined-categories}.

\section{Additional Experiments}
We also conduct experiment on the Hatexplain \citep{mathew2021hatexplain} dataset to show generalization of our method. This dataset is a benchmark for hate speech detection, annotated with explanations provided by human annotators for why certain samples are classified as hateful, offensive, or normal. For training the target model, we focus only on the two classes: Hate and Normal. The groups \textit{Race, Religion, and Gender} are treated as target concepts. We excluded Sexual Harassment due to significant class imbalance. Table \ref{tab:metrics_comparison_hate} shows our results of MeanCG scores computed using Section \ref{sec:cg_compute} scores.

From the results, we observe that for the concept \textit{Race}, correctly classified non-hate samples have a positive CG score, suggesting that reducing the attribution to this concept could lead to misclassification. Similarly, correctly classified hate samples show a positive CG score, indicating that this concept is an essential feature for identifying hate.
The concept \textit{Religion} shows mixed behavior. While correctly classified hate samples have a strong positive CG score, misclassified non-hate samples also display a negative attribution, which indicates over-attribution to this concept in some cases.
The concept \textit{Gender} displays relatively lower attribution scores across all conditions, suggesting that the model relies less on this concept compared to the others. However, the slight positive attribution for misclassified hate samples indicates potential underutilization of this concept.

\begin{table}[h!]
\small  
\centering
\renewcommand{\arraystretch}{1.1}
\setlength\tabcolsep{1.3pt}
\begin{adjustbox}{max width=0.47\textwidth}
\begin{tabular}{>{\centering\arraybackslash}m{1.5cm}>{\centering\arraybackslash}m{1.5cm}>{\centering\arraybackslash}m{1.5cm}>{\centering\arraybackslash}m{1.5cm}>{\centering\arraybackslash}m{1.5cm}}
\toprule
\textbf{Category} & \textbf{Correctly
Classified
(Non-hate)} & \textbf{Correctly
Classified(Hate)} & \textbf{Mis-Classified
(Non-hate)} & \textbf{Mis-Classified
(Hate)} \\
\midrule
Race     & 0.191 & 0.0559   & 0.205  & 0.0132   \\
Religion & -0.0781 & 0.208   & -0.0457 & 0.277   \\
Gender   & -0.0295 & 0.00365   & -0.00756 & 0.0205   \\
\bottomrule
\end{tabular}
\end{adjustbox}
\caption{Comparative CG Scores for Toxicity
Concepts Across Classification Conditions on the HateXplain dataset.}
\label{tab:metrics_comparison_hate}
\end{table}

Table \ref{tab:categories_hate} illustrates the various targeted lexicon sets with the count of corresponding sentences \( S_i^{w_j} \).

\begin{table}[htbp]
\centering
\renewcommand{\arraystretch}{1.4}
\setlength\tabcolsep{1.3pt}
\begin{adjustbox}{max width=0.45\textwidth}
\begin{tabular}{|c|c|}
\hline
\textbf{Targeted Lexicon Sets ($T_i$)} & \textbf{Number of Sentences (\( S_i^{w_j} \))} \\ \hline
Set 1 & 1 \\ \hline
Set 2 & 43 \\ \hline
Set 3 & 460 \\ \hline

\end{tabular}
\end{adjustbox}
\caption{Targeted Lexicon Sets for the Hatexplain Dataset}
\label{tab:categories_hate}
\end{table}

\begin{figure}[h!]
        \centering
    \includegraphics[width=0.45\textwidth]{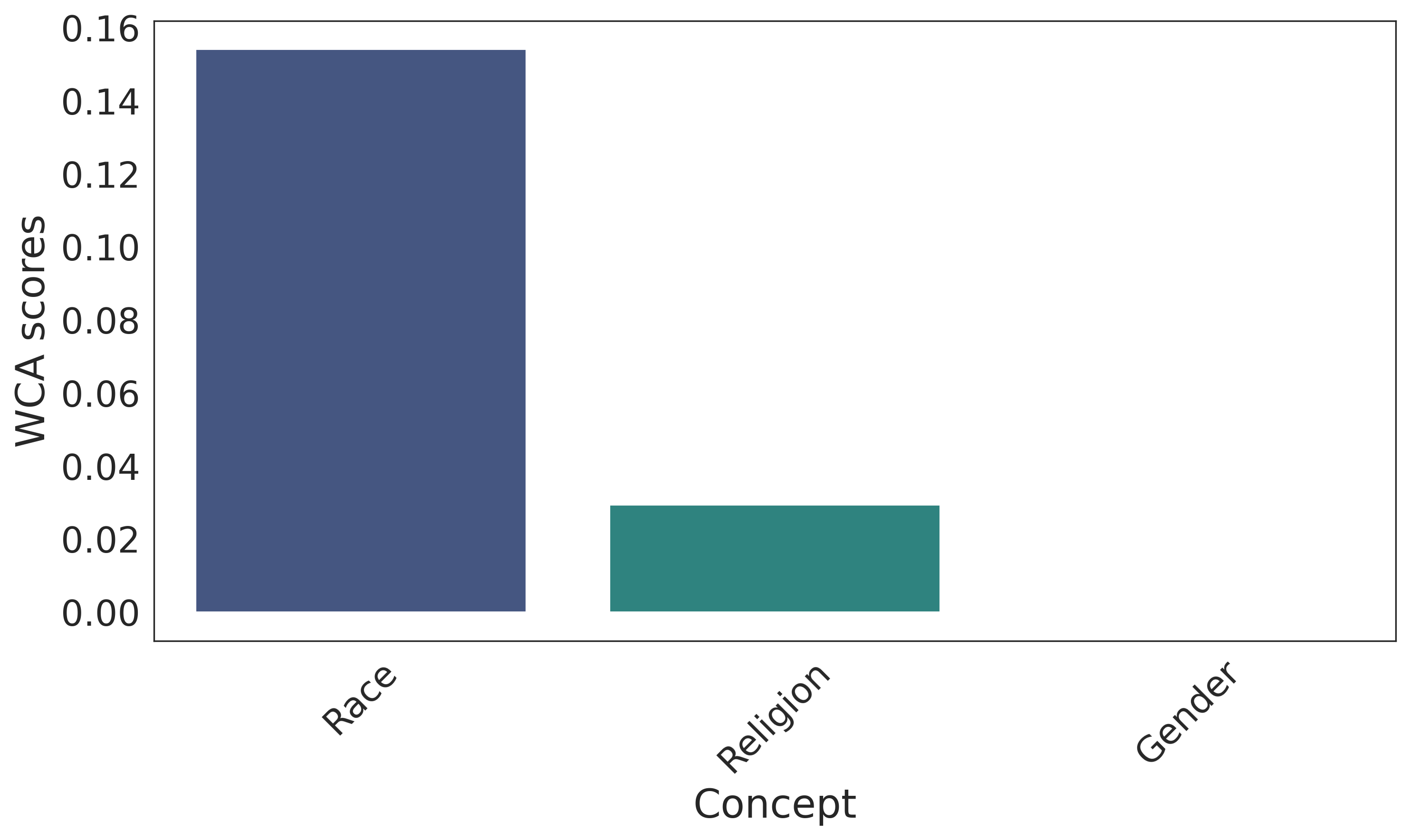}
        \caption{Histogram Set 1 for Hatexplain}
        \label{fig:histogram1_hate}
    \end{figure}

\begin{figure}[h!]
        \centering  \includegraphics[width=0.45\textwidth]{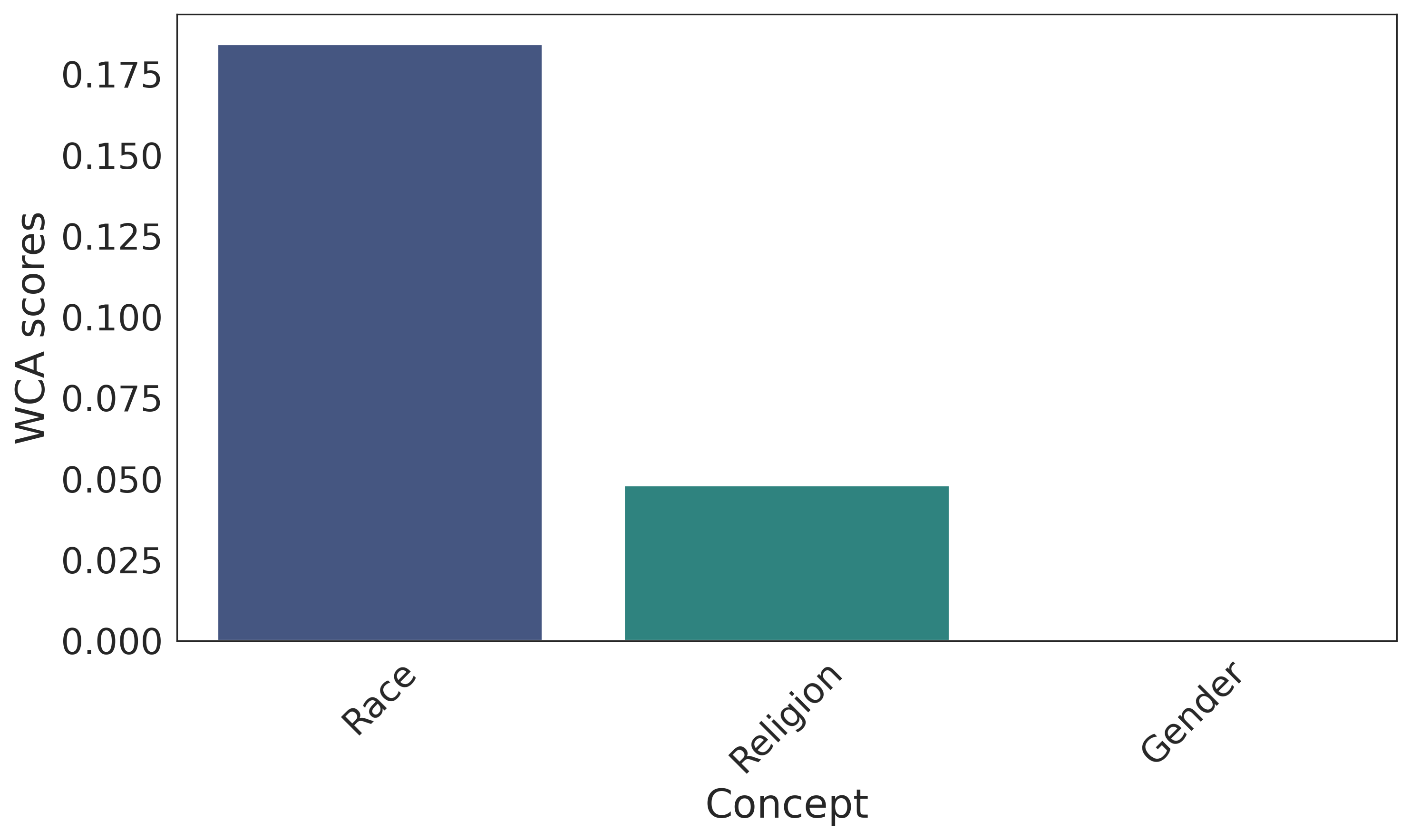}
        \caption{Histogram Set 2 for Hatexplain }
        \label{fig:histogram2_hate}
    \end{figure}

\begin{figure}[h!]
        \centering
        \includegraphics[width=0.45\textwidth]{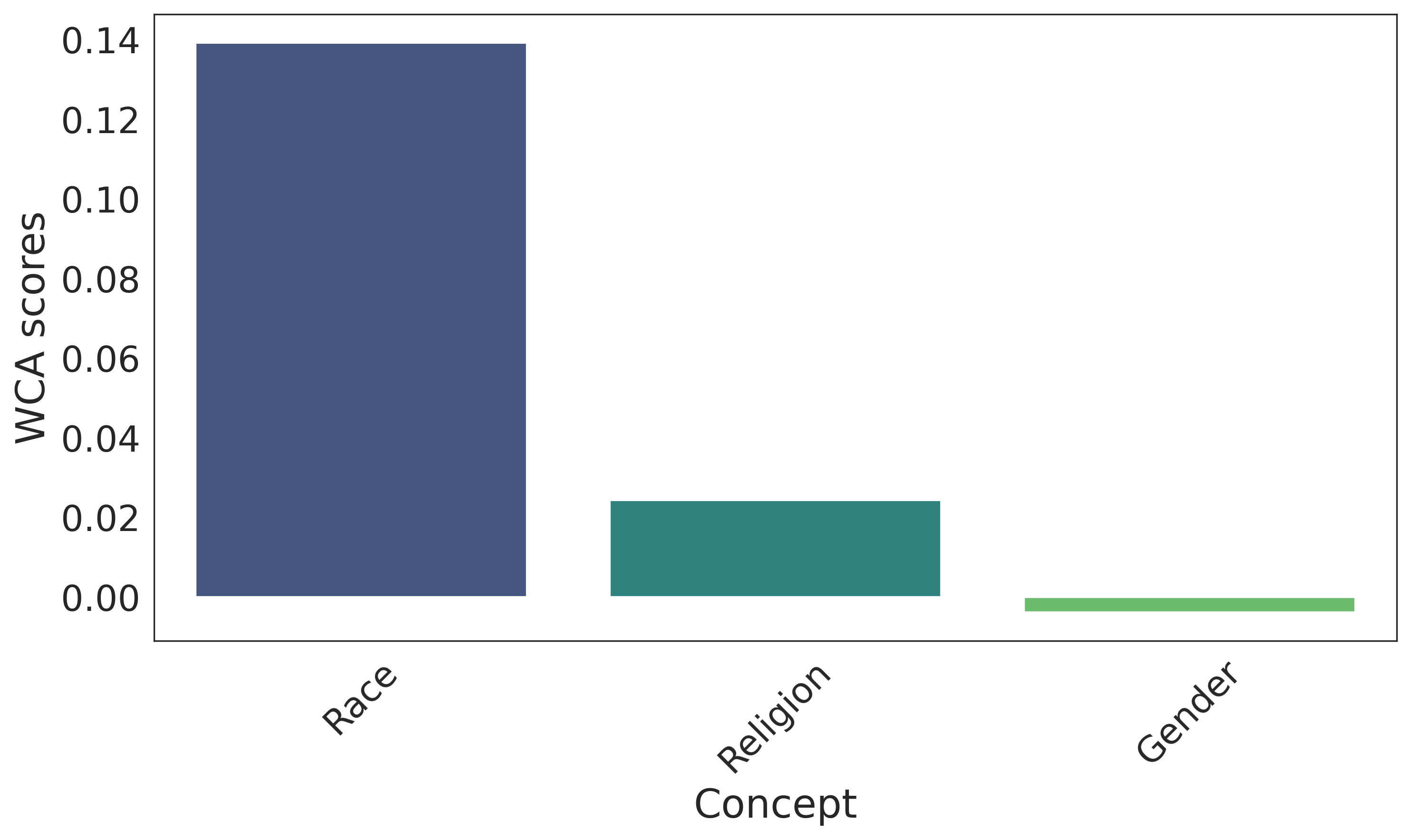}
        \caption{Histogram Set 3 for Hatexplain}
        \label{fig:histogram3_hate}
    \end{figure}

\begin{figure}[h!]
        \centering
    \includegraphics[width=0.45\textwidth]{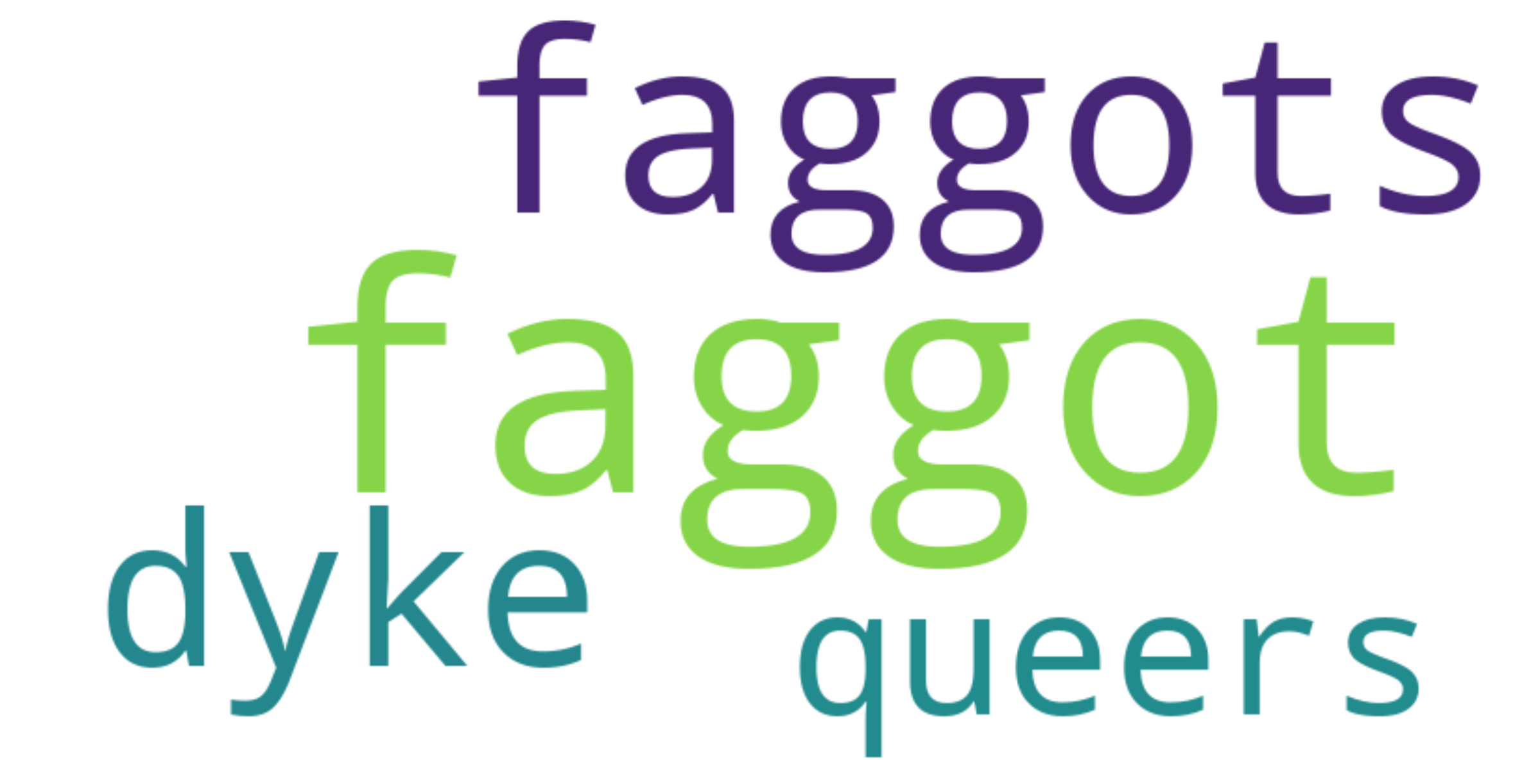}
        \caption{Word Cloud Set 1 for hatexplain}
        \label{fig:word_cloud1_hate}
    \end{figure}

\begin{figure}[h!]
        \centering
        \includegraphics[width=0.45\textwidth]{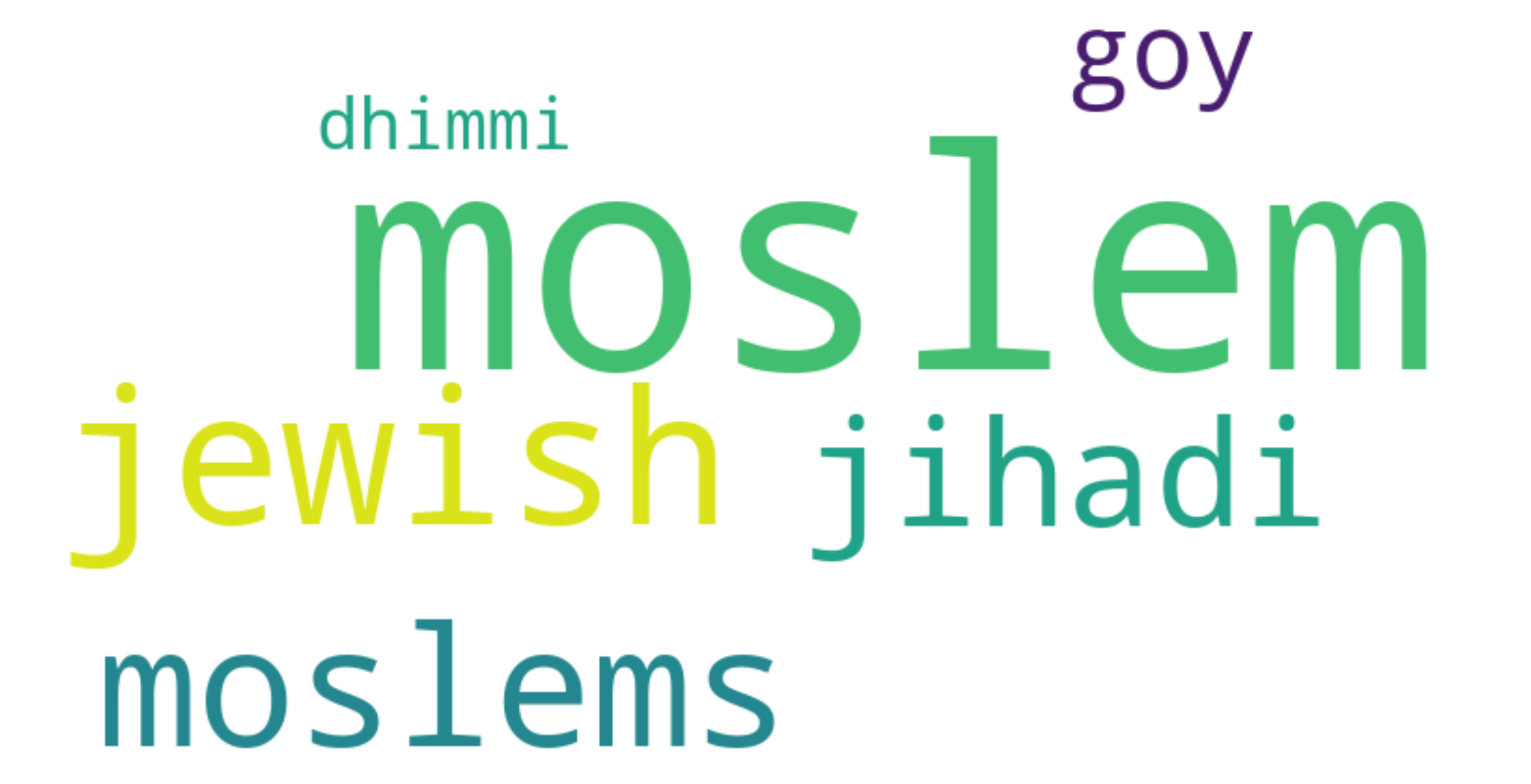}
        \caption{Word Cloud Set 2 for hatexplain}
        \label{fig:word_cloud2_hate}
    \end{figure}

\begin{figure}[h!]
        \centering
        \includegraphics[width=0.45\textwidth]{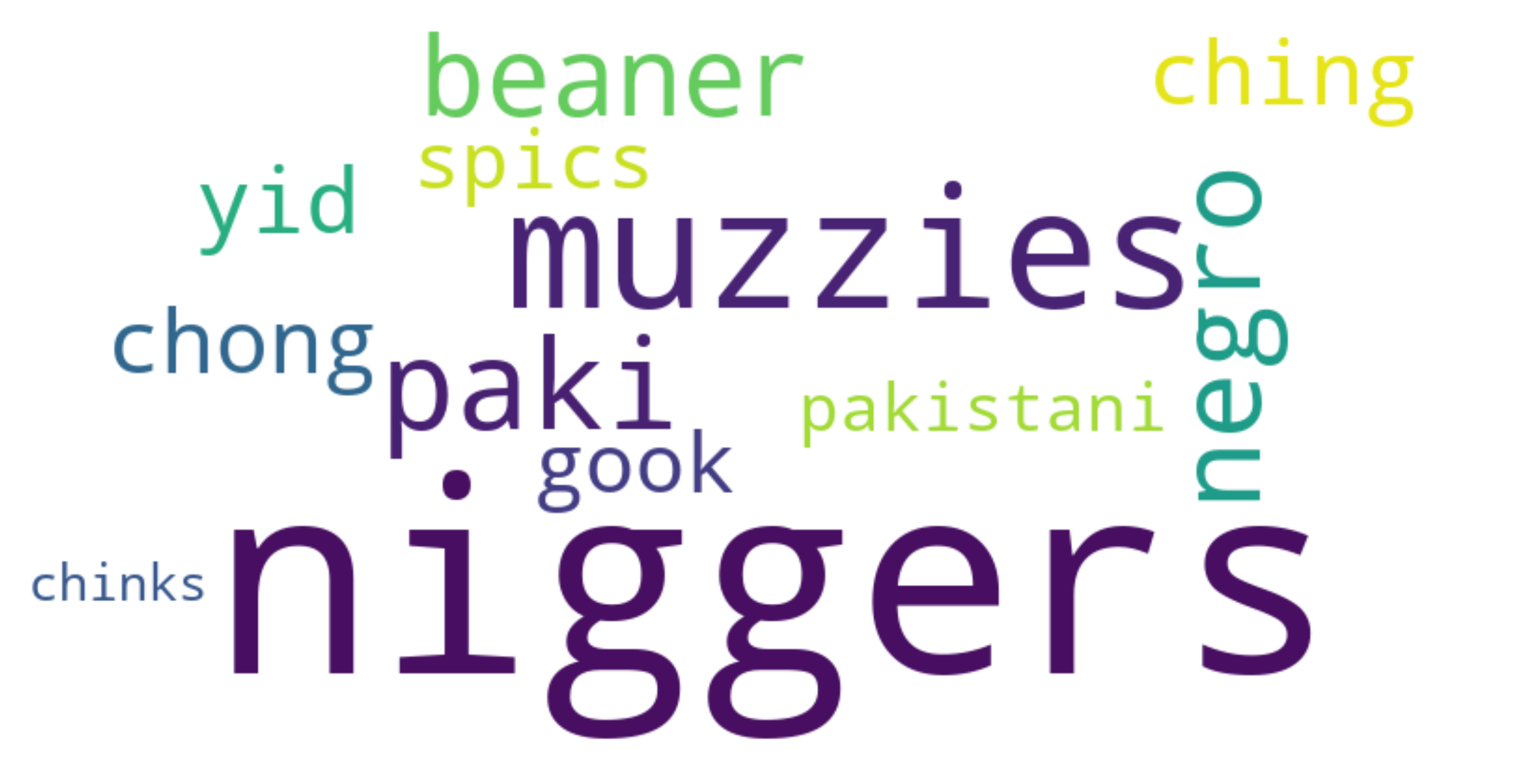}
        \caption{Word Cloud Set 3 for hatexplain}
        \label{fig:word_cloud3_hate}
    \end{figure}

Figure \ref{fig:histogram1_hate}, \ref{fig:histogram2_hate}, \ref{fig:histogram3_hate} and word cloud for the Hatexplain dataset are shown in Figure \ref{fig:word_cloud1_hate}, \ref{fig:word_cloud2_hate}, \ref{fig:word_cloud3_hate}.



\begin{table*}[t!]
\small  
\centering
\renewcommand{\arraystretch}{1.1}
\setlength\tabcolsep{1.3pt}
\begin{adjustbox}{max width=1.0\textwidth}
\begin{tabular}{>{\centering\arraybackslash}m{10.0cm}>{\centering\arraybackslash}m{10.0cm}}
\toprule
\textbf{Category: Sexual Explicit} & \textbf{Category: Threat} \\ \midrule
"And the Quran does not include female genital mutilation. Are you also opposed to male genital mutilation?..." & "You truly have to be a fool to register your weapons in Canada. Only a matter of time before they're all confiscated." \\ \midrule
"Obviously, with female genitalia, he/she would quickly become the most popular 'male' in the prison system." & "He is a bitter fool and knows he has not too much time to be around so, he wants to go as an American hero." \\ \midrule
"Most ignorant thing you have said. No one is 'turned' by sex-ed unless they are already gay..." & "Idiot just took out 2 innocent lives. His family will probably suffer the consequences of his mindless actions." \\ \midrule
"That’s Canada’s PM, the big muddler. But boy, does he bear a striking image at a gay pride parade." & "We can do the same and stifle USA production. We have Europe, Asia and many other countries on our side..." \\ \midrule
"Mass murder, terrorism, perversion, LGBT, gay washrooms, socialism, communism all from this Synagogue of Satan..." & "You are fascinated with sex and don't care about public policy says the man who tied up our entire government..." \\ \midrule
"No kidding. How dare 'Christians' have to live with gay people! How terrible!" & "Lying Trump has been begging the Democrats and the Open Borders Republicans to deliver a DACA Amnesty Bill..." \\ \midrule
"Seems a bit early for a gay pride parade?" & "He drew a RED LINE and ended up kissing ass." \\ \midrule
"I'd scratch it up to your gay friends being as humorless as yourself, but I strongly believe you are making them up..." & "The police know that the whole court system and legal framework is nuts, and that it refuses to put any significant consequences..." \\ \midrule
"In 1968, the evening news was showing, in the southern states, naked black men lynched from trees for being seen with a white woman..." & "This is plain dumb. You want people to intentionally risk their lives for practice?" \\ \midrule
\textbf{Category: Identity Attack} & \textbf{Category: Obscene} \\ \midrule
"(((they))) are the problem, nothing changes. Its tribal. Open your eyes. Its the jews, they are all jews." & "Damn ya'll are ass holes. God forbid you ever make a mistake..." \\ \midrule
"They are parasites trying to ethnically cleanse white people." & "Calling Madeline Albright....you're a dumb arse..." \\ \midrule
"A party with the single message 'We hate whites' is going to be extinguished as a party very soon." & "No, I'm saying, based on previous comments you've made at NCR, that you're a Nazi apologist..." \\ \midrule
"nor will I buy her book... 'blah blah blah white people are bad'" & "MUELLER took my sister hostage with TRACY PRENOSIL AND SARAH DENNIS JENNINGS COWS AND PIGS SATANIC WORSHIPPERS..." \\ \midrule
"(((they))) are the problem, nothing changes. Its tribal. Open your eyes. Its the jews, they are all jews." (Duplicate) & "you are going to be sent to a home for the stupid in 18 months if you keep spewing that stupidity." \\ \midrule
"We used to hang horse thieves." & "This wretched LOSER of a 'woman' - Hillary Clinton - is a self-absorbed, psycho NUT JOB who should be locked up in prison..." \\ \midrule
"Daddy Duggar said incest should be punishable by execution. I'm sure he set up a firing squad for his son." & "What a repulsive psychopath. What an absolute scumbag. This animal needs to be removed before it can do any more damage..." \\ \midrule
"These folks are so narcissistic and stupid it should be a crime for them to reproduce..." & "Couple of them are a bit cartoonish to me. Why the hell is the State screwing around with this?" \\ \midrule
"They are parasites trying to ethnically cleanse white people." (Duplicate) & "And you know this because you read it on fake news sponsored by RT? You poor phony conservatives..." \\ \midrule
"Allison obviously is a tool for the 1\% seeking to defund the social safety net for all Americans who are poor." & "Fing stupid. As a business owner I can refuse service to anyone for any reason..." \\ \midrule
\textbf{Category: Insult} \\
\midrule
"MUELLER took my sister hostage with TRACY PRENOSIL AND SARAH DENNIS JENNINGS COWS AND PIGS SATANIC WORSHIPPERS..." & \\
\midrule
"you are going to be sent to a home for the stupid in 18 months if you keep spewing that stupidity." & \\
\midrule
"This wretched LOSER of a 'woman' - Hillary Clinton - is a self-absorbed, psycho NUT JOB who should be locked up in prison..." & \\
\midrule
"What a repulsive psychopath. What an absolute scumbag. This animal needs to be removed before it can do any more damage..." &  \\
\midrule
"Les stick to the topic. Your idiot in chief is cruel and heartless. He will go to hell." & \\
\midrule
"Ohhh hahahahahahahaha! Another loony fairy tale from the big, fat, orange buffoon." & \\
\midrule
"I find it hard to take this congresswoman clown serious, she looks like a transgendered cowboy/girl..." & \\
\midrule
"Sounds like a loser. Or at least, an angry loser, lol." & \\
\midrule
"He is a complete and utter moron. He adds no value. He must be on vacation again and hence no selfies today." & \\
\midrule
"They are parasites trying to ethnically cleanse white people." & \\
\bottomrule
\end{tabular}
\end{adjustbox}
\caption{Sample Sentences Categorized into Sexual Explicit, Threat, Identity Attack, Obscene, and Insult.}
\label{tab:combined-categories}
\end{table*}

\end{document}